\documentclass[a4paper,onecolumn,oneside,fleqn,10pt]{article}
\usepackage{abstract}
\usepackage{amsmath}
\usepackage{authblk}
\usepackage[english]{babel}
\usepackage{bm}
\usepackage[font={color=blue},figurename=Figure,labelfont={it}]{caption}
\usepackage{cite}
\usepackage{enumitem}   
\usepackage{fchicago} 
\usepackage{float}
\usepackage[left=2cm,right=2cm,top=2cm,bottom=2cm]{geometry}
\usepackage{graphicx}
\usepackage{hyperref}
\usepackage{indentfirst}
\usepackage[utf8]{inputenc}
\usepackage[switch,columnwise]{lineno}
\usepackage{lipsum} 
\usepackage{pdfpages}
\usepackage{rotating}
\usepackage{setspace} 
\usepackage{subcaption}
\usepackage[varg]{txfonts}
\usepackage{wasysym}
\usepackage{fmtcount} 
\usepackage{chngcntr}
\usepackage{titlesec}
\usepackage{listings}
\usepackage[table]{xcolor}  
\usepackage{microtype}
\usepackage{authblk} 
\usepackage{footmisc} 

\counterwithin{equation}{section}
\renewcommand{\theequation}{%
  \thesection.%
  \ifnum\value{equation}<10 0\fi%
  \arabic{equation}%
}

\definecolor{blue}{RGB}{0,0,255}
\definecolor{matlabblue}{rgb}{0,0.4470,0.7410}

\hypersetup{
	colorlinks   = true,        
	linkcolor    = matlabblue,  
	citecolor    = matlabblue,  
	urlcolor     = matlabblue,  
	filecolor    = matlabblue   
	linktoc      = all
}
\DeclareCaptionFont{bluefont}{\color{matlabblue}}
\newcommand{\prior}{{\color{matlabblue}{\textit{prior}\ }}}
\newcommand{\posterior}{{\color{matlabblue}{\textit{posterior}\ }}}

\newcommand{\apriori}{\textit{a priori }}
\newcommand{\aposteriori}{\textit{a posteriori }}

\newcommand{\eg}{\textit{eg. }}
\newcommand{\cf}{\textit{cf. }}
\newcommand{\ie}{\textit{ie. }}

\newcommand{\viceversa}{\textit{vice-versa}}

\usepackage{fancyhdr}
\graphicspath{{images/}{figures/pdf/}}

\counterwithout{equation}{section}       
\renewcommand{\theequation}{\arabic{equation}}

\begin{document}

\title{\huge On the evolution of the concept of probability as a mirror of the evolution of reason \vspace{2cm}}

\author[1]{Le Mouël Jean-Louis\textsuperscript{\dag}}
\author[1]{Courtillot Vincent}
\author[2]{Gibert Dominique}
\author[3,4]{Vladimir Kossobokov}
\author[5]{Boulé Jean-Baptiste}
\author[6]{Zuddas Pierpaolo}
\author[5]{Lopes Fernando}
\author[5]{Marccagi Païkan}
\author[7]{Maineult Alexis}

\affil[1]{\small Académie des Sciences, Institut de France, Paris, France}
\affil[2]{\small  DeepField Sensing, France}
\affil[3]{\small Institute of Earthquake Prediction Theory and Mathematical Geophysics, Russian Academy of Sciences, Moscow, Russia}
\affil[4]{\small Accademia Nazionale delle Scienze detta dei XL, Roma, Italia}
\affil[5]{\small Muséum National d’Histoire Naturelle, CNRS UMR7196, INSERM U1154, Paris, France}
\affil[6]{\small Sorbonne Université, CNRS, METIS,UMR7619, Paris, France}
\affil[7]{\small Laboratoire de Géologie de l’ENS, UMR 8538, Paris, France}

\renewcommand\Authands{ and } 

\makeatletter
\renewcommand\AB@affilsepx{\\[0.5em]}            
\makeatother

\date{}
\maketitle

\begingroup
\renewcommand\thefootnote{\dag}
\footnotetext{In Memoriam Jean-Louis Le Mouël passed away before the finalization of this manuscript. We dedicate this work to his memory.}
\endgroup

\newpage

\begin{abstract}
Over the centuries, probability theory has evolved from a modest calculus of games of chance into a central framework for scientific reasoning under uncertainty. This article argues that probability should not be understood merely as a mathematical tool, but as a historically evolving form of rationality, whose successive transformations reflect deep shifts in the structure of scientific thought itself. From the combinatorial symmetry of Pascal and Fermat to the inductive logic of Bayes and Laplace, from Poisson’s temporalization of events to Kolmogorov’s axiomatic formalization, probability has progressively incorporated uncertainty, time, and coherence into rational judgment. This historical trajectory culminates in modern Bayesian interpretations, exemplified by Tarantola’s conception of probability as a logic of information, in which prior knowledge and observational data are combined through rational inference. While this framework represents a high point in the epistemological maturation of probability, it also reveals its internal limits. Probability theory presupposes well-defined propositions and measurable events; it quantifies uncertainty about facts, but it remains unable to formalize the imprecision of the concepts through which facts are described. The article therefore examines the extension of rationality beyond probability. Fuzzy logic is introduced as a formal response to the problem of vagueness, providing a rigorous language for graded meaning and qualitative judgment. In contrast, the recent rise of deep learning and neural networks is analyzed as a powerful but epistemologically distinct approach. By relying on geometric interpolation and optimization rather than on explicit logical structures, deep learning achieves remarkable predictive performance while bypassing uncertainty representation, conceptual qualification, and causal explanation. By situating probability, fuzzy logic, and deep learning within a unified historical and epistemological perspective, this article clarifies their respective roles and limitations. It argues that contemporary scientific rationality cannot be reduced to data-driven performance alone, but requires the explicit articulation of uncertainty, vagueness, and inference. In this sense, the evolution of probability offers a mirror for the evolution of reason itself, illuminating both the achievements and the unresolved challenges of thinking under uncertainty.

	 
\end{abstract}
\newpage
\section{\label{sec:I}    Introduction} 
Over the centuries, the manner in which probability has been conceived, formalized, and employed has undergone a profound transformation, closely intertwined with the evolution of rational thought itself. For a long time, chance was attributed to fortune, providence, or hidden causes beyond human understanding. In pre-modern societies, randomness was often perceived as a manifestation of divine will or cosmic disorder rather than as an object susceptible to rational analysis (\eg \shortciteNP{daston1992,hald2005,reeves2015}). The gradual emergence of probability theory from the seventeenth century onward marks a decisive intellectual rupture: uncertainty ceased to be merely endured or interpreted symbolically and became instead something to be calculated, reasoned about, and eventually integrated into the very structure of scientific explanation.

The mathematical theory of probability emerged relatively late in the history of ideas, crystallizing in the context of early modern science. Its first systematic formulations arose not from physics or astronomy, but from problems posed by games of chance. The celebrated correspondence between Pascal and Fermat in 1654 concerning the “problem of points” constitutes a founding moment in this history (\cf \shortciteNP{lyraud2003}, p. 407-446 for the french epistolary exchange). What was at stake was not merely the fair division of stakes, but the possibility of subjecting chance itself to rational calculation. From this initial gesture, probability progressively expanded its scope, evolving from a combinatorial arithmetic of equipossible cases into a general framework for reasoning under uncertainty.

Throughout its historical development, probability theory has repeatedly been reformulated, each reformulation reflecting a deeper transformation in the way reason confronts uncertainty. The Bayesian inversion introduced by \shortciteN{bayes1763} and generalized by Laplace (\cf \shortciteNP{laplace1812,laplace1814}) endowed probability with an explicitly inductive and temporal dimension, allowing causes to be inferred from effects and beliefs to be updated in light of new evidence. With \shortciteN{poisson1837}, probability was no longer confined to abstract reasoning but became anchored in empirical reality, giving rise to a genuine dynamics of events and inaugurating modern statistical thinking. The nineteenth and early twentieth centuries further consolidated this trajectory through frequentist interpretations and the discovery of collective regularities such as the law of large numbers and the central limit theorem. Finally, \shortciteN{kolmogorov1933}'s axiomatization endowed probability with full mathematical rigor, completing its formal closure as a measure-theoretic discipline.

Yet probability did not remain a purely mathematical object. Throughout the twentieth century, its epistemological interpretation continued to evolve. Thinkers such as de \shortciteN{finetti1937}, \shortciteN{cox1961} and \shortciteN{jaynes2003} emphasized that probability should be understood as an extension of logic itself, a calculus of rational belief under uncertainty. This perspective finds a particularly clear and operational expression in the work of \shortciteN{tarantola2005}, who interprets probability as a logic of information, especially in the context of inverse problems (\eg \shortciteNP{gibert2024}). In this view, probability is not merely a tool for quantifying randomness or frequencies, but a coherent language for combining prior knowledge and observational data through rational inference. Probability thus appears not simply as a technical apparatus, but as a historically evolving form of rationality.

This article defends the thesis that the evolution of probability theory mirrors the evolution of reason itself. Probability is not reducible to a single interpretation, whether frequentist, subjective, or algorithmic. Rather, it constitutes a conceptual framework that has progressively incorporated symmetry, temporality, induction, empirical confrontation, and logical coherence into scientific reasoning. At the same time, this evolution reveals internal limits. Probability presupposes well-defined propositions and measurable events; it quantifies uncertainty about facts, but it remains silent about the imprecision of the concepts through which facts are described. As contemporary science increasingly confronts complex systems, ill-defined categories, and qualitative judgments, this limitation becomes unavoidable.

Recognizing these limits motivates the exploration of complementary frameworks. In particular, fuzzy logic, introduced by  \shortciteN{zadeh1965} and \shortciteN{zadeh1978}, addresses a distinct dimension of uncertainty: the vagueness inherent in meaning itself. More recently, the rise of artificial intelligence and deep learning has introduced yet another mode of dealing with uncertainty, one that relies on geometric interpolation and optimization rather than on explicit logical structures. While extraordinarily powerful in practice, these methods raise new epistemological questions concerning explanation, causality, and understanding.

The aim of this paper is therefore not merely historical, but philosophical. By tracing the successive transformations of probability, and by situating contemporary computational approaches within this lineage, we seek to clarify what is gained, and what is lost, when reason confronts uncertainty in different ways. Probability, fuzzy logic, and deep learning are not competing tools addressing the same problem; they embody distinct conceptions of rationality, each with its own strengths and limitations.

The structure of the paper reflects this historical and conceptual trajectory. In Section \ref{sec:II}, we examine the birth of probabilistic reasoning in the work of Pascal and Fermat, emphasizing the role of symmetry and combinatorics in the domestication of chance. Section \ref{sec:III} is devoted to Bayes and Laplace, who introduced induction and an explicit arrow of time into probabilistic reasoning, transforming probability into a tool for rational learning. Section \ref{sec:IV} focuses on Poisson and the emergence of probabilistic dynamics, where probability becomes anchored in empirical observation and temporal processes. Section \ref{sec:V} discusses the frequentist turn and the discovery of statistical regularities emerging from disorder, culminating in the law of large numbers and the central limit theorem. Section \ref{sec:VI} addresses the axiomatic closure of probability with Kolmogorov and the resulting epistemic silence of the formal theory. In Section \ref{sec:VII}, we present Tarantola’s interpretation of probability as a logic of information, which represents the most mature epistemological formulation of probabilistic reasoning. Section \ref{sec:VIII} identifies the internal limits of probabilistic expressivity, showing why uncertainty alone is not sufficient to capture all forms of scientific judgment. Section \ref{sec:IX} introduces fuzzy logic as a formal response to the problem of vagueness and graded meaning. Section \ref{sec:X} critically examines deep learning and neural networks, arguing that they constitute a powerful geometric approach that nonetheless bypasses explicit logical reasoning. The conclusion synthesizes these developments and reflects on the future of rationality in the age of uncertainty.

\section{\label{sec:II}   The domestication of chance: symmetry and combinatorics. Pascal, Fermat, and the birth of probabilistic reason}

	The mathematical theory of probability officially emerged in the mid-seventeenth century, when Pascal and de Fermat exchanged letters in 1654 (\eg \shortciteNP{lyraud2003}) concerning a problem of chance that would become famous as the "problem of points". The question was to determine the fair division of stakes in a game interrupted before its conclusion, given the players' respective scores at the moment of interruption. Through this exercise, Pascal and Fermat laid the foundations of probabilistic reasoning by introducing a systematic method to quantify each player’s advantage. Their solution rested on the principle of equipossibility, each future outcome of the game being assumed “equally likely” for the players, and on the concept of mathematical expectation, the mean value of possible gains. This concept of expectation, \ie the sum of possible gains weighted by their respective probabilities, allowed them to define a just division of the stakes. In the simplest version of the problem of points, a game played to three wins, interrupted at a score of 2-1, Pascal explained that two scenarios remained possible, 2-2 or 3-1, each with equal likelihood. The first player was thus guaranteed to recover at least his own stake, $m$ , and had a one-half chance of winning the opponent’s additional stake $m$; hence, he should receive a total of $\dfrac{3m}{2}$, while his opponent would receive $\dfrac{m}{2}$. This reasoning marks the first explicit application of combinatorial calculation to randomness, made possible by tools such as Pascal’s arithmetic triangle, now known as Pascal’s triangle, which provides a systematic way to enumerate favorable and unfavorable cases (\cf \shortciteNP{pascal1665,ore1960}).

This birth of probability marks a major intellectual turning point. For the first time, uncertainty was domesticated through calculation. Whereas earlier civilizations saw in chance the whim of the gods or the workings of fate, Pascal and Fermat proposed to apply reason and logic to games of chance. They formalized, in particular, the rule according to which the probability of an event can be defined as the ratio between the number of favorable cases and the total number of equiprobable cases,
\begin{equation}
	P(E) = \dfrac{\textrm{favorables cases}}{\textrm{possible cases}}.
\end{equation}

This definition, later known as the classical definition of probability, rests on an assumed symmetry, \ie equiprobability, among the conceivable outcomes, reflecting the idea that, in the absence of contrary information, reason postulates an indifference among possible results. Thus, for a fair die, each face has, for example, a $1/6$ chance of appearing, by a purely symmetry-based argument.

It should be noted that this formalization of chance did not occur overnight. The Dutch mathematician \shortciteN{huygens1657} took up the torch, publishing the first treatise on probability, which helped to disseminate these emerging ideas. Yet it was indeed Pascal and Fermat who are credited with initiating the "mathematization of chance", by laying the first foundations of a mathematical theory of the probable. In doing so, they opened the way to a new conception of rationality, \ie the integration of calculation into decision-making under uncertainty. This was a natural extension of the emerging scientific spirit, \ie Galileo, Descartes, and others, which was already seeking rational laws behind natural phenomena; from this point onward, even random phenomena could become the object of rigorous laws and reasoning.

Moreover, Pascal’s approach already contained, in embryonic form, more advanced notions that deserve to be emphasized. In his correspondence, he implicitly employed the idea of conditional expectation to estimate the value of a future stake based on the partial information available, \eg the fact that one player leads 2-1 in an unfinished game. This amounts to calculating the expected gain given the current state of the game, an anticipation of what would later become the notion of conditional probability. This conditional aspect foreshadows one of the fundamental principles of probabilistic reasoning, the updating of an event’s probability when partial information is available. In short, from its very beginnings, the theory of probability appeared as a mirror of reason in action, combining symmetry, \ie equal treatment of possible cases, with conditioning by acquired information.

\begin{figure}[H]
	\centering
	\includegraphics[width=0.7\textwidth]{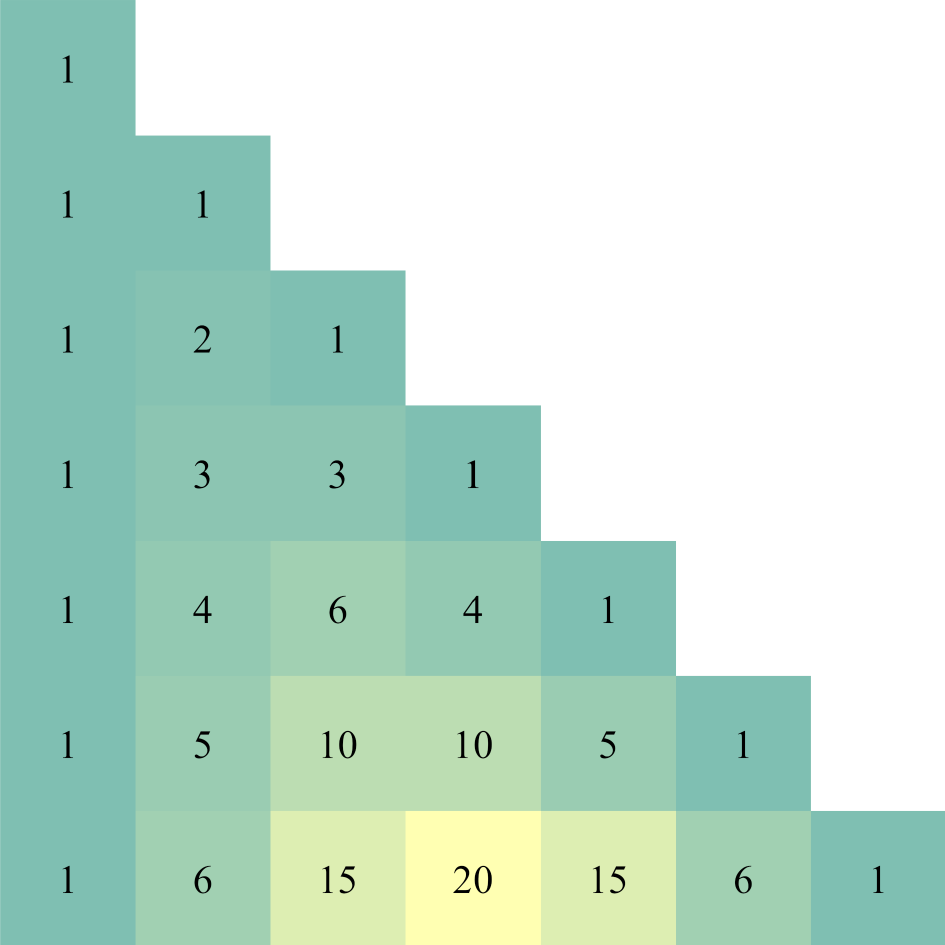}  
	\caption{Pascal’s Triangle. This combinatorial diagram, which organizes the binomial coefficients, reflects one of the founding moments in the history of probability; the transition, initiated in the seventeenth century by Pascal and Fermat, from an arithmetic of games of chance to a genuine mathematical theory of the possible. By systematically enumerating the elementary configurations of a binary event, the triangle makes visible the discrete architecture of randomness and foreshadows the formalization of the calculus of chances, from binomial laws to modern continuous models. It thus plays a key role in the epistemological evolution leading from classical combinatorics to the frequentist interpretation, and ultimately to contemporary Bayesian reinterpretations of random phenomena.}
	\label{Fig:01}
\end{figure}

Figure \ref{Fig:01} provides a visual representation of Pascal’s triangle, as originally employed by Pascal to address early problems in probability. Each cell in the figure corresponds to a binomial coefficient $\left( \begin{array}{c} n \\ k \end{array}\right)$ , that is, the number of distinct ways to obtain exactly $k$ occurrences of a given outcome, here heads, which we designate as the event of interest, over  successive tosses of a fair coin. The triangular structure directly displays the combinatorial progression, the first row contains a single ‘1’, the second two ‘1’s, the third ‘1-2-1’, and so forth. The color scale, ranging from pale green to lighter yellow in the central region, highlights the rapid increase in the number of configurations when $k$ lies near $n/2$ corresponding to the largest number of distinct sequences yielding an intermediate number of heads. Conversely, the edges of the triangle, in a deeper green, remind us that there is only one possible sequence that produces either zero heads or $n$ heads: always obtaining the same face.

This simple geometric construction encapsulates the core intuition of the early calculus of chances as it emerged in the seventeenth century. Suppose a fair coin is tossed $n$ times. The $2^n$ possible sequences of heads and tails are all assumed to be equally likely, in accordance with Pascal’s principle of indifference, in the absence of further information, reason must treat all possible outcomes as symmetrical. The $n^{th}$ row of Pascal’s triangle shows how these $2^{n}$ sequences are distributed according to the number of heads observed. For $n=6$, illustrated in the Figure \ref{Fig:01}, one reads the sequence ‘1-6-15-20-15-6-1’, this means that there are 20 sequences containing exactly three heads, but only 6 sequences containing one or five heads, and a single sequence containing zero or six heads. Dividing each of these counts by $2^6 = 64$ immediately yields the binomial distribution with parameter $p=1/2$, which describes the exact distribution of the number of heads obtained in six independent trials. This figure highlights two essential ideas for understanding how Pascal, followed by Fermat, transformed chance into an object of rational analysis. First, it shows that the apparent ‘disorder’ of successive coin tosses conceals an underlying combinatorial structure that is perfectly regular. Second, it reveals that this structure is symmetric; the number of sequences yielding $k$ heads is equal to the number yielding $n-k$ heads. This symmetry plays a fundamental role in the emergence of the modern notion of equiprobability, in the early games of chance studied by Pascal, nothing allowed one to favor one sequence over another; reason therefore compels us to treat all cases alike, to enumerate them, and to compute probabilities by comparing each count to the total.

Thus, Pascal’s triangle is not merely a mnemonic device for arranging binomial coefficients, it stands as one of the earliest visual matrices of probabilistic thought. It reveals how a random situation can be dissected into a finite set of possibilities, all equally plausible, whose examination leads to a quantitative measure of uncertainty. In this sense, the Figure \ref{Fig:01} materializes the historical transition traced in the present article, the moment when seventeenth-century French rationality, heir to Descartes and contemporary with Pascal, begins to calculate chance, reducing uncertainty to a geometry of possibilities. Beneath the arithmetic of the triangle lies the outline of modern probability, understood as the ratio between the combinatorial structure of the world and the impartiality of reason toward all possible cases.

\section{\label{sec:III}  Induction and the arrow of time. Bayes and Laplace: probability as rational learning}
	After the combinatorial foundations of the seventeenth century, the eighteenth century witnessed a major conceptual expansion; probability became a tool for inductive inference. Two emblematic figures illustrate this intellectual revolution. First, \shortciteN{bayes1763} proposed to infer the probability of a cause from observed effects, the so-called problem of inverse probability. Second, \shortciteN{laplace1774}, who independently rediscovered Bayes’s result and made it the cornerstone of a general theory of scientific induction. 

Bayes’s theorem states that for two events $A$ and $B$, one has,
\begin{equation}
	P(A|B) = \dfrac{P(B|A)P(A)}{P(B)}.
\end{equation}

In other words, the probability of $A$ given $B$ is obtained by multiplying the prior probability of $A$ by the likelihood of observing $B$ under the assumption $A$, and by normalizing this product by the marginal probability of $B$. Bayes presented this theorem in the context of an urn problem, where one seeks to estimate an unknown probability from observed trials, in modern terms, he was computing a \posterior probability of a parameter by combining \prior information with experimental data. Although Bayes himself could not fully develop all the implications of his formula, his 1763 paper, published posthumously by his friend Price, laid the foundations of Bayesian inference. From that moment onward, it became possible to "reason backward" from effects to causes in a rational way, by quantifying how new observations update our degrees of belief. In this sense, Bayes introduced a genuine arrow of time into probabilistic reasoning, one starts from an initial state of knowledge, \ie the \prior probability, and refines it progressively as new data arrive, \ie the \posterior probability. This temporal orientation, from the past, or \prior hypotheses, toward the future, or updated conclusions, constitutes a major conceptual innovation. It turns probability into a dynamic tool, no longer a mere static symmetry of idealized cases.

Laplace, often regarded as the true father of probabilistic statistics, went further by generalizing and systematically applying Bayes’s method. In his \textit{Théorie analytique des probabilités} (\cf \shortciteNP{laplace1812}) and \textit{Essai philosophique sur les Probabilités} (\cf \shortciteNP{laplace1814}), Laplace developed a unified vision of probability, at once a combinatorial calculus, inherited from Pascal, and a universal inductive method. He applied probabilistic reasoning to a wide range of fields; astronomy with measurement errors, determination of orbits, etc., demography, insurance, and even jurisprudence with assessment of the reliability of testimonies, among others. It was Laplace who popularized Bayes’s famous formula in its modern form and interpreted it in terms of probable causes and effects. In a sense, he transformed Bayes’s formula into a principle of reasoning: “to infer causes from observed effects”.

Laplace is also known for his principle of indifference, also called the principle of insufficient reason, which generalizes the idea of equiprobability; in the absence of any information, all possible outcomes should be assumed equally likely. For example, we can read in \shortciteN{laplace1812}, 2sd book, chapter 1, p. 178, $\S$2 : "\textit{The theory of probability consists in reducing all events that may occur under given circumstances to a certain number of equally possible cases, that is, cases about whose occurrence we are equally undecided, and in determining, among these cases, the number of those that are favorable to the event whose probability is sought}". This principle allowed him to assign \apriori objective probabilities even to unique or non-repeatable events. For instance, Laplace calculated the probability that a scientific discovery was due to chance, or that an astronomical phenomenon would recur. A famous example is Laplace’s rule of succession; if an event has occurred n times in succession without exception, \eg the sunrise observed each morning, then the probability that it will occur again next time is estimated as,
\begin{equation}
	P_{next} = \dfrac{n+1}{n+2}
\end{equation}

Using this formula, \shortciteN{laplace1814} obtained a probability extremely close to 1 ($\approx 0.99999945$), that the Sun would rise again the following day, given that it had always risen on all known days of human history. This example, seemingly paradoxical at first glance, illustrates how Laplace’s probabilistic reasoning treats an uncertain future on the basis of observed past events;  although absolute certainty remains unattainable, the computed probability provides a rational guide, almost a practical certainty in this case, for anticipating future outcomes. One clearly sees here the role of the arrow of time; probabilistic reasoning combines the past, \ie the $n$ occurrences already observed, with a symmetry principle, \ie no \prior bias, hence the $\dfrac{1}{n+2}$ as an initial weight of ignorance, to project a quantitative measure of confidence into the future; \ie the $(n+1)-th$ occurrence.

Laplace famously encapsulated this philosophy in a striking sentence: “\textit{From this Essay, one sees that the theory of probabilities is, at bottom, nothing more than common sense reduced to calculation.}” (\cf \shortciteNP{laplace1814}, p. 190, $\S$2). For him, the calculus of probabilities provided a quantitative formalization of what the prudent human mind already does intuitively, weighting its beliefs according to the evidence. This remark shows that, as early as the beginning of the nineteenth century, probability was perceived as a mirror of sound reasoning under uncertainty, that is, an extension of common sense capable of achieving a level of precision and coherence that intuition alone could not guarantee. Laplace further added that the theory of probability “\textit{It enables us to assess with accuracy what reasonable minds grasp by a kind of instinct, without often being able to account for it.}” (\cf \shortciteNP{laplace1814}, p. 190, $\S$2). In other words, it renders explicit and refines our intuitive judgments of reason.

It is worth noting, somewhat ironically, that Laplace remained a convinced determinist at the philosophical level, his famous “Laplace’s demon” an imaginary intellect that could predict the future if it knew all positions and velocities at a given instant, stands as the emblem of this view: \textit{“An intellect which, at a given instant, knew all the forces that animate nature and the respective positions of the beings that compose it, and which were moreover vast enough to submit these data to analysis, would embrace in a single formula the movements of the greatest bodies in the universe and those of the lightest atom: nothing would be uncertain for it, and the future, as well as the past, would be present to its eyes}” (\cf \shortciteNP{laplace1814}, pp. 4-5).  For Laplace, chance was merely an appearance born of our ignorance. Yet it is precisely to compensate for this unavoidable ignorance that probability becomes indispensable;  in the absence of perfect knowledge of the world, we employ probabilistic calculation as a substitute for complete understanding. This tension between determinism and probability is characteristic of the era; classical reason held that uncertainty could, in principle, be eliminated through full information, yet in practice it had to acknowledge the necessity of the probable in order to act without such complete information. The contribution of Bayes and Laplace was to provide the intellectual tools to rationalize these wagers on the unknown.

\begin{figure}[H]
	\centering
	\includegraphics[width=1\textwidth]{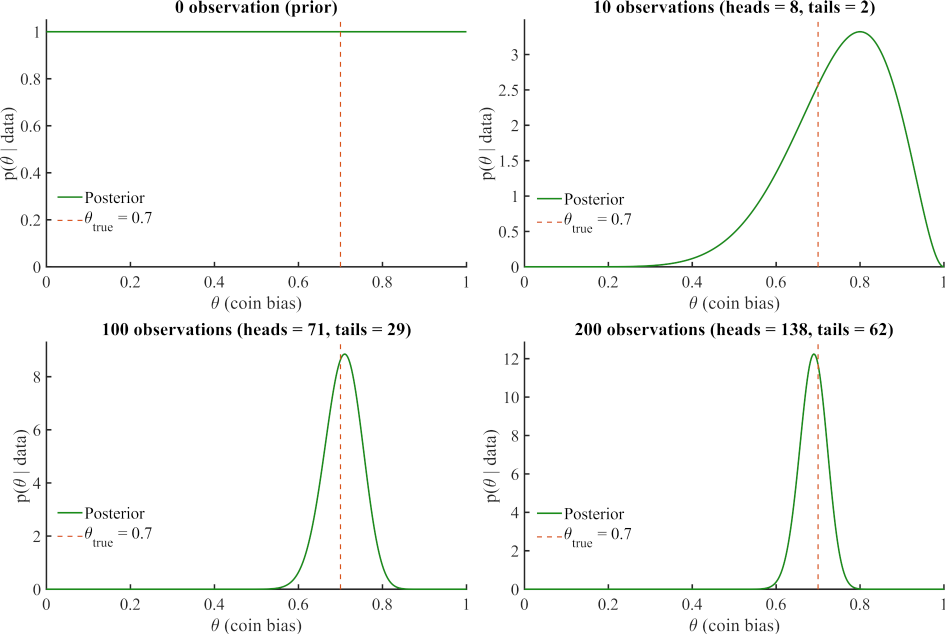}  
	\caption{Bayesian inference process applied to estimating the bias of a coin whose true probability of ‘heads’ is $\theta = 0.7$. The uniform prior (absence of observation) represents a state of complete ignorance. After 10 tosses, the likelihood begins to shape the belief, though uncertainty remains significant. With 100 and then 200 observations, the posterior distribution becomes increasingly narrow and centered around the correct value. This sequence illustrates the essential dynamic of the Bayesian approach; a cumulative, coherent, and quantified learning process in which knowledge is not imposed by the data but continuously revised by them. It also highlights the way in which the notion of probability evolves from a purely combinatorial framework (\cf. Figure \ref{Fig:01}) toward an inferential framework based on the rational updating of beliefs.}
	\label{Fig:02}
\end{figure}

Figure \ref{Fig:02} illustrates, using a simple coin-toss example, how a Bayesian line of reasoning allows one to move from an initial state of ignorance to an increasingly precise estimate of an unknown parameter, in this case, the bias of the coin. We assume that a coin is tossed repeatedly, with heads coded as the outcome of interest (value 1) and tails as 0. The parameter $\theta$ denotes the unknown probability of obtaining heads on each toss. To fix ideas, the data shown in the figure were generated using a ‘true’ value $\theta_{true} = 0.7$, indicated by the red vertical dashed line. The observer, however, does not know this value: it must be inferred from the successive outcomes of the tosses.

In summary, with Bayes and Laplace, probability acquired an inductive and temporal dimension. No longer was it merely a matter of enumerating \apriori symmetric configurations; one now learns from \aposteriori data. The concept of conditional probability and Bayes’s formula introduce a directionality into reasoning; we move from past correlations, \ie observed data and initial hypotheses, toward a future updating of our beliefs, \ie the \posterior. The language of probability calculus thus becomes the language of experimental reason;  with each new observation, reason revises its conclusions coherently. This idea of a continuous adaptation of beliefs under the influence of new information represents a genuine epistemological arrow of time, the hallmark of the Enlightenment’s recognition, in the eighteenth century, that uncertainty is not a failure of knowledge, but an intrinsic component of it.

\section{\label{sec:IV}   From expectation to events. Poisson and the emergence of probabilistic dynamics}
Within the great lineage of thinkers who shaped the theory of probability, Poisson occupies a singular place,  at once the heir of Laplace and a precursor to modern developments in statistics and stochastic processes. Poisson was both a student and an admirer of Laplace, whom he readily described as his master, and he extended Laplace’s work by seeking to anchor probability in measurable reality, beyond its purely combinatorial foundations. His name remains attached to a fundamental law, the Poisson distribution,  introduced in his 1837 treatise "\textit{Recherches sur la probabilité des jugements en matière criminelle et en matière civile}", yet his influence extends far beyond that formula. In a sense, \shortciteN{poisson1837} transformed probability into a dynamics of chance; he gave substance to the idea that random events are not merely to be counted, but can be conceived as a continuous temporal process endowed with its own statistical regularities.

In his treatise, \shortciteN{poisson1837} set out to solve a concrete problem; how to evaluate, through calculation, the reliability of verdicts delivered by criminal juries. Following in the footsteps of \shortciteN{condorcet1785} and \shortciteN{laplace1812}, he sought to apply probabilistic reasoning to human decisions, while for the first time incorporating empirical data drawn from official French criminal statistics (1825-1835). The undertaking was audacious, \shortciteN{poisson1837} aimed to quantify the probability of judicial error as a function of the average competence of jurors and the observed rate of convictions, thereby producing a kind of moral and rational assessment of French justice. He called this endeavor “measuring the moral state of the country”. This intellectual gesture marks a turning point; probability ceased to be merely an art of hypothetical reasoning and became an instrument of social observation, grounded in data. Through this attempt to synthesize theoretical calculation and empirical measurement, Poisson inaugurated what could be described as mathematical statistics.

But posterity would remember above all from this treatise the unveiling of a new law of chance, the Poisson law, which formalizes the probability that a rare event occurs a certain number of times within a given interval. \shortciteN{poisson1837} began with the binomial law of \shortciteN{bernoulli1713} and \shortciteN{laplace1812} and derived from it a remarkable limit; when the number of trials  becomes very large and the probability of success $p$ becomes very small, so that $np = \lambda$, remains constant, the distribution of occurrences tends toward,
\begin{equation}
	P(k) = \dfrac{e^{-\lambda} \lambda^{k}}{k!}.
\end{equation}

This law, sometimes called the “law of small numbers”, elegantly captures the regularity of rare phenomena, whether the number of deaths from a specific cause, the number of accidents occurring in a single day, or, later, the arrival of telephone calls at a switchboard. Beneath this simple formula lies a profound intuition: even chance itself, when observed across large ensembles, obeys stable regularities. The apparent disorder of individual events resolves, through aggregation, into a lasting statistical structure. This idea, that the multiplicity of contingencies gives rise to order, stands as one of the guiding threads of modern probabilistic thought and one of Poisson’s most fertile intellectual legacies.

\begin{figure}[H]
	\centering
	\includegraphics[width=1\textwidth]{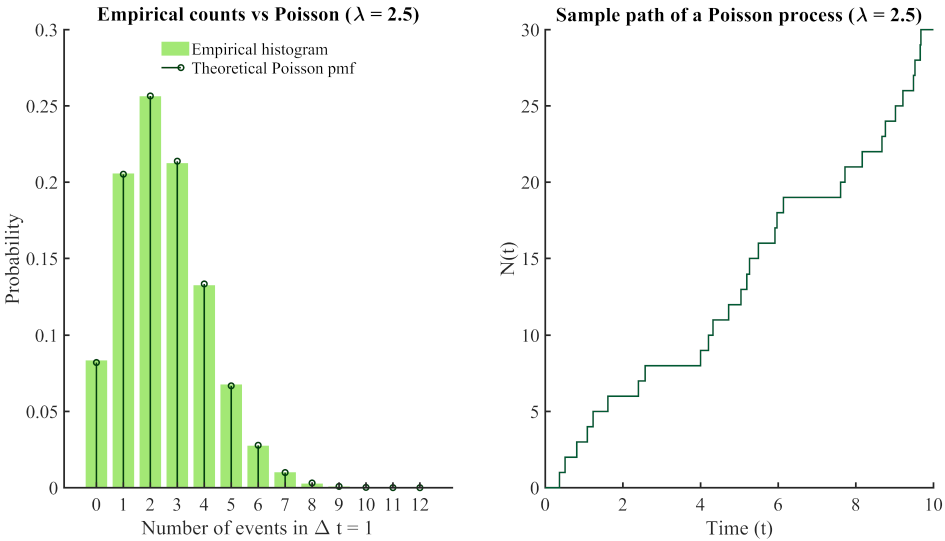}  
	\caption{Empirical manifestation of the Poisson law, the “law of small numbers” introduced by Poisson (1837). The left panel compares the empirical distribution of the number of occurrences observed within a unit interval $\Delta t = 1$ to the theoretical Poisson probability mass function $P(k) = \dfrac{e^{-\lambda} \lambda^{k}}{k!}$ with $\lambda = 2.5$. Although each event is rare and independent, their aggregated counts exhibit a remarkable stability, a property that Poisson derived as a limit of the binomial law of Bernoulli (1713) and Laplace (1812) when the number of trials becomes large and the success probability small. The right panel shows a sample path of the corresponding Poisson counting process: a staircase-like trajectory whose jumps occur at unpredictable times yet accumulate at an average rate that remains strikingly regular. This dual perspective, distributional and temporal, embodies Poisson’s central insight; that even phenomena governed by chance alone can form enduring statistical structures when considered over large ensembles. The law of small numbers thus illustrates the emergence of order from contingency, one of the foundational intuitions of modern probabilistic thought.}
	\label{Fig:04}
\end{figure}

Figure \ref{Fig:04} offers a concrete and intuitive illustration of the statistical structure captured by Poisson’s “law of small numbers”. The left panel shows the empirical distribution of the number of events occurring in a fixed interval of length $\Delta t = 1$, obtained from repeated simulations of a rare-event mechanism with average rate $\lambda = 2.5$. Although each trial is governed by chance and nothing guarantees that the same number of events should appear from one realization to the next, the histogram rapidly stabilizes around the theoretical Poisson distribution $P(k) = \dfrac{e^{-\lambda} \lambda^{k}}{k!}$ represented by the black markers. This agreement exemplifies the remarkable regularity that Poisson (1837) uncovered when he derived his law as the limiting form of the binomial distribution studied earlier by \shortciteN{bernoulli1713} and \shortciteN{laplace1812}; when the number of trials becomes very large while the probability of success becomes very small, the distribution of the total number of occurrences settles into a universal and analytically tractable shape. The right panel complements this distributional perspective by showing a sample path of the associated Poisson counting process $N(t)$. The trajectory evolves in irregular jumps separated by random waiting times, capturing the unpredictability of each individual occurrence, yet the overall growth of the process reflects the stable average rate $\lambda$. The path thus reveals, in temporal form, the same epistemological message conveyed by the histogram; beneath the apparent disorder of isolated events lies a persistent statistical order, emerging from the aggregation of many independent contingencies. Whether we examine the counts over a fixed interval or the accumulation of events through time, the Poisson law demonstrates how the study of rare phenomena gave rise to one of the foundational insights of modern probabilistic reasoning: the idea that randomness, when viewed across large ensembles, obeys precise and reproducible laws.

Beyond the law itself, Poisson implicitly introduced a new notion, that of the Poisson process, a temporal flow of independent random events occurring at a constant average rate. Without formulating it with modern rigor, Poisson was already conceiving of chance as a function of time, rather than as a mere distribution of states. He assumed that events occur independently of one another, in disjoint intervals, and that their number over a given duration follows the law he had discovered. This vision, remarkably modern, anticipated an entire branch of twentieth-century stochastic theory, from the counting processes of \shortciteN{galton1814} and \shortciteN{yule1925} to the foundations of queueing theory and statistical physics. By establishing a connection between probability, frequency, and time, Poisson ushered the theory of chance into a kinematics of events; the question was no longer only how many times an outcome occurs, but how and at what rate it unfolds over time. In this lies, in embryonic form, the birth of the modern theory of stochastic processes.

Poisson’s work also exemplifies the nineteenth century’s characteristic tension between mathematical reason and social reality. In seeking to apply the calculus of probability to the administration of justice, he extended the French rationalist tradition, the lineage from Pascal to Laplace that sought to subject chance to reason, but at the same time revealed its practical limits. His contemporaries, in both the legal and scientific worlds, received his work with caution. Applying the calculus of the probable to moral or judicial decisions appeared to them both bold and misplaced; as with Condorcet before him, society was not yet ready to accept that mathematical reason could pass judgment on human affairs. Yet behind these resistances, Poisson raised an essential question: what is the true scope of probabilistic reasoning? Can the uncertainty of human judgment really be quantified ? Is probability confined to the science of the external world, or can it be extended to the domains of decision, ethics, and society ? These questions, which would reemerge a century later in the debates on applied statistics and artificial intelligence, show how deeply Poisson anticipated the epistemological challenges of modern modeling.

\shortciteN{poisson1837} must finally be situated within the historical dialogue between \shortciteN{bayes1763} and \shortciteN{laplace1812}. While Laplace had given the calculus of probability its philosophical and universal dimension, that of a general method of rational induction, Poisson ensured its descent into the concrete. Where Laplace established the principles of probabilistic reasoning and proposed ideal applications (\eg celestial mechanics, the theory of errors, hypothetical jury votes), Poisson turned toward the actual measurement of phenomena, the use of available statistical data, and the construction of quantitative indicators of society. In this sense, he was a forerunner of the frequentist perspective; the goal was no longer merely to calculate \apriori probabilities, but to estimate them from observed frequencies, following an empirical inductive logic. His approach, without breaking with that of Laplace, both complemented and extended it; he introduced into the theory a dimension of observation and confrontation with the real world. If Bayes and Laplace had made the calculus of probability an instrument of reasoning, Poisson made it an instrument of measurement as well.

\section{\label{sec:V}    Regularity from disorder. Frequencies, large numbers, and statistical determinism}
Throughout the nineteenth and early twentieth centuries, the concept of probability continued to grow in scope and precision, in parallel with advances in science and in the philosophy of science. Two complementary trends characterized this period; on the one hand, the desire to relate probability to observed frequencies, \ie the so-called frequentist or objective view; and on the other hand, the pursuit of axiomatic rigor aimed at establishing probability theory as an autonomous mathematical discipline.

\shortciteN{bernoulli1713} had paved the way for frequentism with his law of large numbers. This theorem, published posthumously in \textit{Ars Conjectandi}, established that the observed frequency of an event tends to approach its theoretical probability as the number of trials becomes very large. More precisely, Bernoulli showed that if an experiment is repeated a large number $N$ f times, the proportion of observed successes converges, in a probabilistic sense, toward $p$ the probability of success in each trial. In doing so, Bernoulli expressed for the first time a formal link between mathematical probability and its empirical representation in the real world. This result had a profound conceptual impact; it justified the estimation of probabilities through experimental statistics and reinforced the idea that the laws of chance can be verified through the accumulation of data, something far from evident \apriori.

In the nineteenth century, this frequentist interpretation gradually took root. Philosophers and mathematicians such as \shortciteN{venn1866} and \shortciteN{borel1909} emphasized that “probability has meaning only as the long-run limit of frequency”. Probability was then defined as the value toward which the frequency of occurrence of an event tends when the number of trials approaches infinity. In this view, time, or at least the number of repetitions,  plays an essential role; it is by letting $N$ grow that probability reveals itself. 

Frequentist reasoning thus rests on an objectivist conception; probability is seen as a property of the real world, a stable frequency, that can be estimated with increasing accuracy through large series of observations. This perspective complements the preceding Bayesian one;  whereas Bayes and Laplace regarded probability as the degree of belief of a rational observer, \ie a rather subjective viewpoint, even though Laplace considered it universal, frequentism views it as a measurable property of repetitive phenomena. In practice, the two often converge, stable frequencies justify \apriori intuitions, and \viceversa, but philosophically, a shift occurs in probabilistic reasoning by the end of the nineteenth century, toward a more empirical outlook.

At the same time, major mathematical advances came to consolidate the theory. The development of error calculus and mathematical statistics, around figures such as \shortciteN{gauss1809} and \shortciteN{legendre1806}, integrated probability into the analysis of scientific data. \shortciteN{gauss1809} introduced the famous normal law, or Gaussian distribution, to describe the distribution of measurement errors. A few years earlier, \shortciteN{moivre1718} and \shortciteN{laplace1812} had discovered the central limit theorem, showing that the sum of many independent random effects tends toward a Gaussian distribution. This probabilistic explanation of the ubiquity of the “bell curve” in natural phenomena provided further evidence that randomness obeys regular laws when a large number of variables are considered. The idea that statistical order emerges from individual disorder reinforced confidence in probabilistic reasoning as an integral part of scientific thought, scientific reason came to acknowledge that, even without strict determinism, there exists a collective determinism, a regularity in the large numbers, that probability allows us to grasp.

\begin{figure}[H]
	\centering
	\includegraphics[width=1\textwidth]{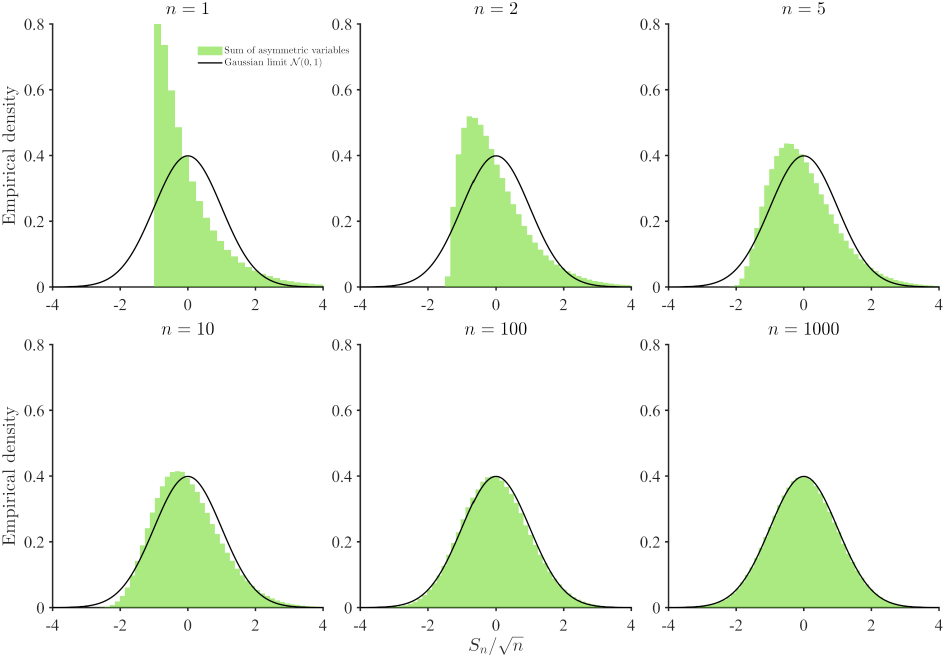}  
	\caption{Illustration of the Central Limit Theorem, as formulated by de Moivre (1718) and Laplace (1812). The panels display the empirical distribution of the normalized sum $S_{n}/n$ of $n$ independent and strongly skewed random variables (green). For small $n$, the distribution is far from Gaussian, but as the number of components increases, from $n=1$ to $n=1000$ , the shape stabilizes and converges toward the limiting Gaussian law $\mathcal{N}(0,1)$ (black curve). This visual demonstration reflects a foundational idea in the history of probabilistic reasoning; even when individual causes are irregular or asymmetric, their collective effect exhibits a universal and highly regular structure. In this sense, the “bell curve” emerges not from determinism, but from the statistical order produced by large numbers.}
	\label{Fig:03}
\end{figure}

Figure \ref{Fig:03} offers a detailed and vivid illustration of the central idea formulated by \shortciteN{moivre1718} and later extended by \shortciteN{laplace1812}; that the combined effect of many independent random contributions tends toward a universal and highly regular form. In each panel, we observe the empirical distribution of the normalized sum $S_{n}/\sqrt{n}$ of $n$ independent random variables drawn from a markedly asymmetric exponential distribution. The choice of such a skewed base distribution is deliberate; it emphasizes that nothing in the microscopic behavior of the individual terms resembles the Gaussian curve, nor even hints at symmetry. For $n=1$, the distribution is entirely dominated by this asymmetry, presenting the familiar long tail of the exponential. For $n=2$ and $n=5$, the asymmetry persists, though it becomes softened by the simple act of aggregation. As $n$ increases to 10, 100, and ultimately 1000, a striking transformation takes place. The distribution becomes progressively smoother and more symmetric, its peak stabilizes, and its tails adjust in a way that draws the entire histogram closer to the superimposed Gaussian curve $\mathcal{N}(0,1)$. This visual convergence is not merely a numerical curiosity; it embodies a profound epistemological shift in the understanding of randomness. de Moivre and Laplace recognized that even when individual events are governed by irregular, unpredictable fluctuations, the aggregated behavior exhibits a remarkable regularity; one that is stable, universal, and mathematically quantifiable. The Figure \ref{Fig:03} thus dramatizes one of the foundational insights of probabilistic thinking since the Enlightenment: order emerges from disorder when the number of contributing variables becomes large. This emergence is not deterministic in the classical sense, for the micro-level remains governed by chance; yet the collective produces a form of determinism at the level of ensembles, what Laplace himself would later describe as a “remarkable regularity” intrinsic to large numbers. The “bell curve”, apparently ubiquitous across biological, physical, and social phenomena, finds its explanation not in the structure of the underlying causes, but in the combinatorial laws that govern their accumulation. By showing a highly skewed distribution spontaneously organizing itself into a Gaussian profile through mere summation, Figure \ref{Fig:03} embodies this conceptual leap; the realization that probability does not merely quantify uncertainty but reveals a deep structural tendency of nature. It shows that randomness, when considered in isolation, is chaotic, but when viewed collectively, obeys regular laws that allow prediction, inference, and scientific explanation. In this sense, the central limit theorem is not just a technical result; it is a cornerstone of the modern scientific worldview, reconciling the absence of strict determinism at the microscopic level with the emergence of robust statistical determinism at the macroscopic scale.

Despite its many successes, by the end of the nineteenth century probability theory still suffered from an ambiguous epistemological status. Many mathematicians regarded it as insufficiently rigorous, resting on poorly defined notions: what exactly is “equiprobability” or “randomness” outside the context of finite combinatorial cases? This is why the twentieth century marked the completion of a crucial stage: the axiomatic formalization of probability.

\section{\label{sec:VI}   Axiomatization and epistemic silence. Kolmogorov and the mathematical closure of probability}
\shortciteN{kolmogorov1933} published Foundations of the Theory of Probability, in which he proposed a rigorous axiomatic framework inspired by measure theory. Kolmogorov defined a probability $P$ as a measure, in the sense of Lebesgue measure theory, on a sample space $\Omega$ satisfying three simple axioms 1) $P(A) \geq 0$ for every event $A$, 2) $P(\Omega) = 1$ and 3) for any countable family of disjoint events $A_i$, $P(\bigcup_{i}A_{i}) = \sum_{i} P(A_{i})$. Kolmogorov’s axioms gave probability the same degree of rigor as geometry or algebra. In particular, the entire theory follows logically from these postulates, with notions such as conditional probability and independence becoming derived definitions rather than primitive concepts.

Kolmogorov’s contribution was to establish probability as a fully fledged mathematical discipline, independent of the need for philosophical interpretation in order to be coherent. From that point onward, one could say, “no matter what chance is in itself, mathematical probability is a consistent and well-defined tool”. This axiomatization legitimized probability in the eyes of all mathematicians. It was this stage that legitimized probability as a mathematical discipline in its own right” (\eg \shortciteNP{barone1978}). It also responded to a program envisaged by Hilbert, his sixth problem, which aimed to axiomatize physics and all branches of applied mathematics. After \shortciteN{kolmogorov1933}, probability theory could thus be taught in the same way as topology or algebra, without concern for earlier semantic paradoxes, such as Bertrand’s paradox of geometric probabilities, now elegantly resolved through the framework of Lebesgue measure.

However, the philosophical interpretation of probability continued to evolve throughout the twentieth century. An intense debate persisted between advocates of an objective view of probability, ie. as frequency or physical propensity, and those defending a subjective view, \ie as degree of belief. The statistician Fisher and the frequentist school (\eg \shortciteNP{neyman1933,neyman1977}) dominated the first half of the twentieth century in practice, rejecting the use of Bayesian \apriori assumptions and favoring methods based solely on observed frequencies, such as hypothesis testing and confidence intervals. Conversely, thinkers such as de \shortciteN{finetti1937} in Italy and \shortciteN{ramsey1931} in England defended, in the 1930s, a radically subjective interpretation; “\textit{Probability does not exist},” de Finetti claimed, it is merely the way a coherent individual bets on an event. De Finetti even provided a behavioral characterization of probability as the fair betting rate of a rational agent, leading to the idea that the coherence of bets enforces the usual laws of probability. This subjectivist interpretation thus rejoined the Bayesian conception by a different path; it is rational to apply Bayes’s theorem to update one’s beliefs after new observations, and although the measure of these beliefs is personal, it remains bound by universal constraints of coherence, those expressed by \shortciteN{kolmogorov1933}’s axioms, or equivalently by \shortciteN{cox1961}’s axioms of inductive logic.

The mid-twentieth century also witnessed a reversal of trends, with a revival of the Bayesian approach led by figures such as \shortciteN{jeffreys1939} argued that “\textit{Bayes’s theorem stands to probability as Pythagoras’s theorem does to geometry}” (\eg \shortciteNP{savage1954,edwards1963,lindley1972}). In particular, \shortciteN{jaynes2003} formulated the idea that probability theory is an extension of classical logic to situations of uncertainty, in other words, a generalized logical calculus in which “true” and “false” are replaced by degrees of plausibility between 0 and 1. This perspective provides a mature synthesis of the idea that probability is a mirror of reason reasoning under uncertainty: just as deductive logic reflects the structure of certain reasoning, probability reflects the structure of uncertain reasoning.

From a scientific standpoint, the second half of the twentieth century and the beginning of the twenty-first have seen probability theory permeate virtually every field: quantum physics (where probability is intrinsic to the laws of nature, breaking Laplacian determinism), Earth sciences, biology, social sciences, economics, and of course computer science and artificial intelligence. The concept of the arrow of time has even become central in statistical physics, where the increase of entropy, defined in probabilistic terms, explains the irreversibility of macroscopic phenomena.

Contemporary scientific reason thus accepts a form of fundamental indeterminism, \eg in quantum mechanics, the outcome of a measurement is inherently probabilistic, while using probability to produce the most precise predictions possible. This natural incorporation of randomness into the very core of our understanding of the world may well represent the culmination of the evolution of reason itself; what was once perceived as a deficiency of knowledge, \ie an ignorance to be overcome, is now recognized as an irreducible feature of reality, one that knowledge must integrate. In this sense, probability theory has become as indispensable to modern science as differential calculus once was to classical mechanics.

Thus, at the dawn of the twenty-first century, probability stands simultaneously as a rigorous mathematical theory, a methodological foundation for empirical science, and a component of the philosophy of knowledge. Its historical evolution,  from games of chance to machine-learning algorithms, mirrors the paradigm shifts in the way human reason approaches uncertainty.

\section{\label{sec:VII}  Probability as a logic of information. Tarantola and the epistemology of inference}
	As we have emphasized, a modern and unifying view considers probability theory as an extended logic for reasoning under uncertainty. This view is exemplified in the work of Tarantola, who played a major role in introducing Bayesian methods into inverse problems,  that is, the inference of causes from observed effects, for example determining the internal structure of the Earth from seismic recordings. Tarantola adopts a remarkably pure epistemological perspective; in a scientific problem where one seeks to estimate unknown parameters from measured data, all unknown quantities must be modeled as random variables representing our uncertainty. Rather than searching for a single deterministic solution to an inverse problem, often ill-posed and non-unique, he proposes to characterize the entire set of possible solutions through a posterior probability density (\eg \shortciteNP{tarantola1982}).

\shortciteN{tarantola2005}’s approach is fundamentally Bayesian. One starts from a \prior density $P(p)$, representing the initial information about the unknown parameters $p$, \eg a probability distribution expressing an initial estimate of the geological structure; then one models the measurement process through a conditional data density, or likelihood $P(d|p)$, giving the probability of obtaining the data $d$ for a given choice of parameters $p$; finally, one applies \shortciteN{bayes1763}’s formula to obtain the posterior density $P(p|d)  \propto P(d|p) P(p)$, which constitutes the complete solution to the inverse problem. Tarantola emphasizes that this combination of prior information, arising for instance, from physical knowledge or an initial model, with the information contained in new data is analogous to the logical operation “AND”; both pieces of information must be satisfied simultaneously. Mathematically, the fusion of these two independent sources of knowledge is achieved by a product of densities, or more generally a convolution in certain continuous cases, just as in Boolean logic conjunction requires that two conditions be true at the same time. Tarantola formalized this analogy by requiring that the rule for combining information obey the same properties as logical conjunction, transposed into probabilistic terms. This perspective amounts to saying: “\textit{Reasoning with probabilities is reasoning almost as in logic, but allowing for intermediate degrees of truth}”.

\shortciteN{tarantola2005}’s approach illustrates how contemporary probabilistic reasoning operates; all sources of uncertainty, measurement errors, natural variability, and model imprecision, are integrated into a single coherent framework, and the rules of probability (\ie Bayes and related principles) are then used to deduce the final state of knowledge once the data have been taken into account. This approach is now common not only in geophysics but also across many fields of engineering and science. For example, in robotics and artificial intelligence, one speaks of a Bayes filter, of which the well-known \shortciteN{kalman1960} filter is a linear-Gaussian case (\eg \shortciteNP{chen2003}); an algorithm that updates, in real time, the probability distribution of a system’s state as new observations arrive. Once again, the idea is that the machine’s knowledge at any given time is represented by a probability distribution over possible states, which is combined, via a likelihood computation, often equivalent to a convolution with a transition model followed by multiplication by an observation density and normalization, with incoming data to yield the updated knowledge. The parallel with Tarantola’s framework is clear. In this sense, Tarantola extends Laplace’s vision: “common sense reduced to calculation” becomes, in Tarantola’s formulation, “logic reduced to probabilistic calculation.”

It is interesting to note that Tarantola, like Jaynes and de Finetti before him, recognized that the rigor of probabilistic calculation is precisely what guarantees the global coherence of inductive reasoning. Any other ad hoc way of combining information would risk violating rational principles, \eg by ignoring relevant evidence or by counting it twice. By using probability as a language, one automatically inherits the additive and multiplicative coherence enforced by Kolmogorov’s axioms. In a sense, one can say that reason has now integrated chance: what once belonged to intuition or instinct, such as giving greater weight to more precise information, is now codified by computation. An observation with low uncertainty yields a sharply peaked likelihood and therefore carries greater weight in Bayesian updating than a noisy observation with a diffuse likelihood.

Tarantola summarizes this by emphasizing the necessity of accounting for all possible states of information; the solution to a problem is no longer a single number or model, but the entire collection of possible models, each weighted by its posterior probability. This epistemological dimension of Tarantola's work was later emphasized by Mosegaard (\cf \shortciteNP{mosegaard2011}), who described Tarantola's legacy as a quest for consistency, symmetry, and simplicity. In this sense, the probabilistic formulation of inverse problems is not merely a computational technique, but the expression of a broader rational requirement: all pieces of information entering an inference problem must be combined according to rules that are coherent, invariant with respect to arbitrary choices of parametrization, and as simple as possible.

In this way, he echoes the well-known statement that “correlation is not causation”. Indeed, in a purely frequentist or descriptive approach, finding a strong correlation between two phenomena is not sufficient to infer a causal link. The Bayesian-logical framework complements this insight by providing a structure for testing alternative causal hypotheses and determining which is most probable in light of the data, that is, by enabling probabilistic causal inference.

\begin{figure}[H]
	\centering
	\includegraphics[width=1\textwidth]{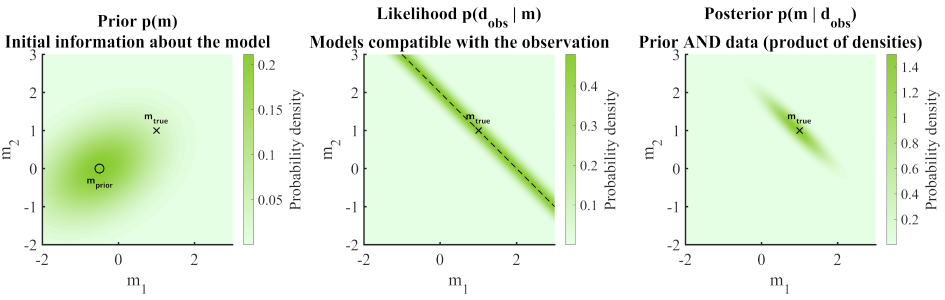}  
	\caption{Conceptual demonstration of Bayesian inference in a two-dimensional parameter space, following the epistemological interpretation introduced by Tarantola. In the left panel, the prior distribution $p(m)$ represents the initial state of knowledge about the model parameters ($m_1,m_2$); the density is centered on an \apriori estimate $m_{prior}$ reflecting information available before any observation is made. The middle panel shows the likelihood function $p(d_{obs} | m)$, which quantifies the compatibility between the data and each possible model. Its elongated shape reflects the fact that many parameter combinations fit the observation equally well, forming a “manifold of acceptable models”. The right panel displays the posterior distribution $p(m | d_{obs})$ obtained through the Bayesian update. Mathematically, this update is the product of the prior and likelihood densities; conceptually, it corresponds to a logical conjunction (“AND”): the posterior contains only those models that satisfy both the initial knowledge and the observational evidence. The resulting density is narrower, more informative, and sharply peaked around the true parameter value $m_{true}$. This figure exemplifies the central Bayesian idea that scientific inference does not replace prior knowledge with data, but rather fuses the two into a coherent, quantitatively updated state of belief, a synthesis that reflects how scientific reasoning integrates experience with established understanding.}
	\label{Fig:05}
\end{figure}

Figure \ref{Fig:05} illustrates, in a two-dimensional parameter space, the full logic of Bayesian inference as interpreted by Tarantola; not as a mechanical update rule, but as a fusion of independent sources of information. The left panel shows the prior distribution $p(m)$, which expresses what is known about the parameters ($m_1,m_2$) before any observation is taken into account. The density is centered near an \apriori estimate $m_{prior}$, shown by the open circle, and spreads broadly across the model space, reflecting initial uncertainty. The true model $m_{true}$ represented by a cross, lies outside the region of maximal prior probability, emphasizing that prior knowledge alone is incomplete or biased. The middle panel displays the likelihood function $p(d_{obs} | m)$ , which evaluates how compatible each model is with the observed data. Its geometry is striking; rather than isolating a single parameter value, the likelihood forms a narrow, elongated ridge, a set of models that all reproduce the observation equally well. This ridge highlights an essential aspect of inverse problems; data typically constrain only certain combinations of parameters, leaving entire directions of uncertainty unconstrained. The dashed line represents the axis of this degeneracy, passing close to the true model $m_{true}$. The right panel shows the posterior distribution $p(m | d_{obs})$ which emerges from the Bayesian synthesis of these two informational components. Mathematically, the \posterior is the product of prior and likelihood; epistemologically, it is the region of the model space where the two sources of knowledge overlap. As Tarantola emphasized, the Bayesian update is not a replacement of \prior knowledge by data but a logical conjunction; the \posterior contains only those models that satisfy both the \prior information and the observational constraint. The resulting density is sharply focused near $m_{true}$ demonstrating that the fusion of incomplete but complementary information leads to a far more precise characterization of the parameters than either source alone. Thus, the figure makes visually explicit a central idea of modern probabilistic reasoning; knowledge in science does not arise from data alone, nor from theory alone, but from the coherent integration of the two. Through this fusion, uncertainty is reduced in a principled way, and the true model becomes identifiable within a structured landscape of probabilities.

\section{\label{sec:VIII} When uncertainty is not enough. The limits of probabilistic expressivity}
The historical trajectory traced so far suggests that probability theory has progressively extended the scope of rationality by providing ever more refined tools for reasoning under uncertainty. From the combinatorial symmetry of Pascal (\cf section \ref{sec:II}) to the inductive dynamics of Bayes and Laplace (\cf section \ref{sec:III}), from Poisson’s (\cf section \ref{sec:IV}) temporalization of events to Kolmogorov’s axiomatic closure (\cf section \ref{sec:VI}), and finally to Tarantola’s interpretation of probability as a logic of information (\cf section \ref{sec:VII}), probability has gradually acquired the status of a universal language for scientific inference. In its most mature form, probabilistic reasoning appears capable of integrating prior knowledge, observational data, model uncertainty, and measurement noise into a single coherent framework. One might therefore be tempted to conclude that probability, properly understood, exhausts the rational treatment of uncertainty.

Yet this conclusion proves premature when one examines more closely the conditions under which probabilistic reasoning operates. Probability theory presupposes that the objects of reasoning are well defined. Events must be identifiable, hypotheses must be clearly formulated, and the space of possible outcomes must be specified in advance, even if only implicitly. Uncertainty, in the probabilistic sense, concerns the truth value of propositions whose meaning is already fixed. One asks whether a given hypothesis is true or false, whether a parameter takes one value rather than another, or whether an event occurs or not, and probability assigns degrees of confidence to these alternatives. In all these cases, the indeterminacy lies in our knowledge of the world, not in the definition of the concepts themselves.

However, many scientific situations do not conform to this idealized structure. In a wide range of domains, the difficulty is not merely to decide whether a well-posed hypothesis is true, but to determine what the hypothesis actually means. The problem is no longer confined to uncertainty about outcomes, but extends to imprecision in the very categories used to describe them. Scientific practice abounds in statements such as “the signal is weak,” “the model is acceptable,” “the structure is compatible with the data,” “the anomaly is significant,” or “the system is close to equilibrium.” These judgments are neither purely subjective nor strictly binary. They rely on graded assessments, contextual interpretation, and qualitative distinctions that resist sharp boundaries. In such cases, asking for the probability of an event presupposes that the event has already been crisply delineated, which is precisely what is at stake.

This limitation becomes particularly evident in inverse problems and complex systems, where Tarantola’s probabilistic framework otherwise proves remarkably powerful. Even when a posterior distribution has been rigorously constructed, the interpretation of its structure often requires additional layers of judgment. Which regions of parameter space should be regarded as “plausible,” “acceptable,” or “physically meaningful”? At what point does a model cease to be “compatible” with the data? Such questions cannot be answered by probability values alone, because they involve thresholds and categories whose definition is not dictated by probability theory itself. The posterior density quantifies degrees of belief, but it does not, by itself, specify how these degrees should be mapped onto qualitative decisions or linguistic descriptions.

This observation reveals an internal boundary of probabilistic expressivity. Probability excels at quantifying uncertainty about well-defined propositions, but it remains silent about the vagueness of the propositions themselves. It assumes that the language in which hypotheses are formulated is already precise, whereas in practice, scientific reasoning often operates at the interface between quantitative data and qualitative concepts. The difficulty is not accidental; it reflects a structural feature of probability theory. As a calculus of measures, probability requires measurable sets, that is, sharply defined subsets of a sample space. Vagueness, by contrast, concerns situations where membership itself is a matter of degree, where an object can belong partially to a category without fully satisfying its defining criteria.

Historically, this tension has been obscured by the success of probabilistic methods in domains where concepts could be idealized and boundaries drawn with sufficient clarity, such as games of chance, physical measurements, or repeated experiments under controlled conditions. As scientific inquiry increasingly turns toward complex, heterogeneous, and poorly delimited phenomena, however, the limits of this idealization become apparent. In such contexts, uncertainty is inseparable from imprecision, and reasoning requires tools capable of handling both simultaneously. Probability alone cannot adjudicate questions whose formulation already involves graded notions and continuous transitions between categories.

Recognizing this limitation does not amount to a rejection of probability theory. On the contrary, it is precisely because probability has achieved such a high degree of coherence and universality that its boundaries can now be clearly identified. The issue is not that probabilistic reasoning is flawed, but that it is incomplete when confronted with forms of uncertainty that arise from the indeterminacy of meaning rather than from the randomness of outcomes. This realization prepares the ground for an extension of rationality beyond probability, one that seeks to formalize not only uncertainty about facts, but also the vagueness of the concepts through which facts are apprehended.

In this sense, the question that now arises is no longer “what is the probability that a given event occurs?” but rather “to what extent does a given situation belong to a given conceptual category?” Addressing this question requires a different, though complementary, formal language. The historical evolution of rationality thus appears to demand a new step, analogous to that which led from combinatorics to induction, and from induction to axiomatic probability. It is at this juncture that fuzzy logic enters the scene, not as a competitor to probability, but as a response to a distinct and irreducible dimension of uncertainty: the imprecision inherent in meaning itself.

\section{\label{sec:IX}   Beyond probability: fuzzy logic and the formalization of vagueness}
The limitation identified in the previous section calls for a formal extension of rationality capable of addressing not uncertainty about facts, but imprecision in the meaning of the concepts used to describe them. This extension was proposed in the second half of the twentieth century by \shortciteN{zadeh1965} and \shortciteN{zadeh1978}, through the introduction of fuzzy sets and fuzzy logic. Whereas probability theory assigns degrees of belief to well-defined propositions, fuzzy logic assigns degrees of membership to categories whose boundaries are intrinsically gradual. The two frameworks thus address distinct, though complementary, dimensions of uncertainty.

In fuzzy logic, a statement such as “$x$ belongs to the set $A$” is no longer treated as either true or false. Instead, membership is quantified by a function $\mu_A(x)$ taking values between 0 and 1, which expresses to what extent $x$ satisfies the defining properties of $A$. This formalism captures a mode of reasoning that is ubiquitous in scientific practice and natural language, where concepts such as “large,” “stable,” “near,” or “acceptable” do not admit sharp thresholds. Importantly, this is not a matter of ignorance or incomplete information; it reflects the structure of the concepts themselves. Vagueness is not noise to be eliminated, but a constitutive feature of many epistemic categories.

From this perspective, fuzzy logic does not compete with probability theory, nor does it aim to replace it. Probability expresses a degree of confidence in a hypothesis whose meaning is fixed; fuzzy logic expresses a degree of compatibility between a situation and a concept whose meaning is inherently graded. The two logics operate at different levels. Probability quantifies uncertainty about truth, while fuzzy logic formalizes imprecision of meaning. Their domains overlap in practice, but their epistemological roles are distinct. A statement may be highly probable while remaining conceptually vague, or conceptually precise while probabilistically uncertain.

This distinction becomes particularly salient in complex systems and inverse problems, where probabilistic reasoning often reaches its expressive limit. Even when a posterior distribution is rigorously constructed, the interpretation of its structure frequently relies on qualitative judgments: which models should be regarded as plausible, which regions of parameter space are acceptable, and which solutions are clearly inadmissible. These judgments implicitly invoke graded categories that are not encoded in probability densities themselves. Fuzzy logic makes this implicit layer explicit by providing a formal language for such qualitative assessments.

An illustrative example is provided by the comparison between classical probabilistic simulated annealing and a fuzzy-logic variant applied to an optimization problem. In the probabilistic scheme, candidate solutions are accepted or rejected according to a stochastic criterion, allowing even clearly suboptimal configurations to be explored with non-zero probability. In the fuzzy scheme, candidate solutions are evaluated through graded linguistic categories such as “bad,” “medium,” or “good,” which act as qualitative filters. The resulting behavior is not merely a matter of computational efficiency; it reflects a different epistemic stance. Random exploration is no longer unconstrained, but modulated by explicit judgments of coherence and plausibility. The algorithm thus embodies, in formal terms, a mode of reasoning closer to human evaluative judgment.

\begin{figure}[H]
	\centering
	\includegraphics[width=1\textwidth]{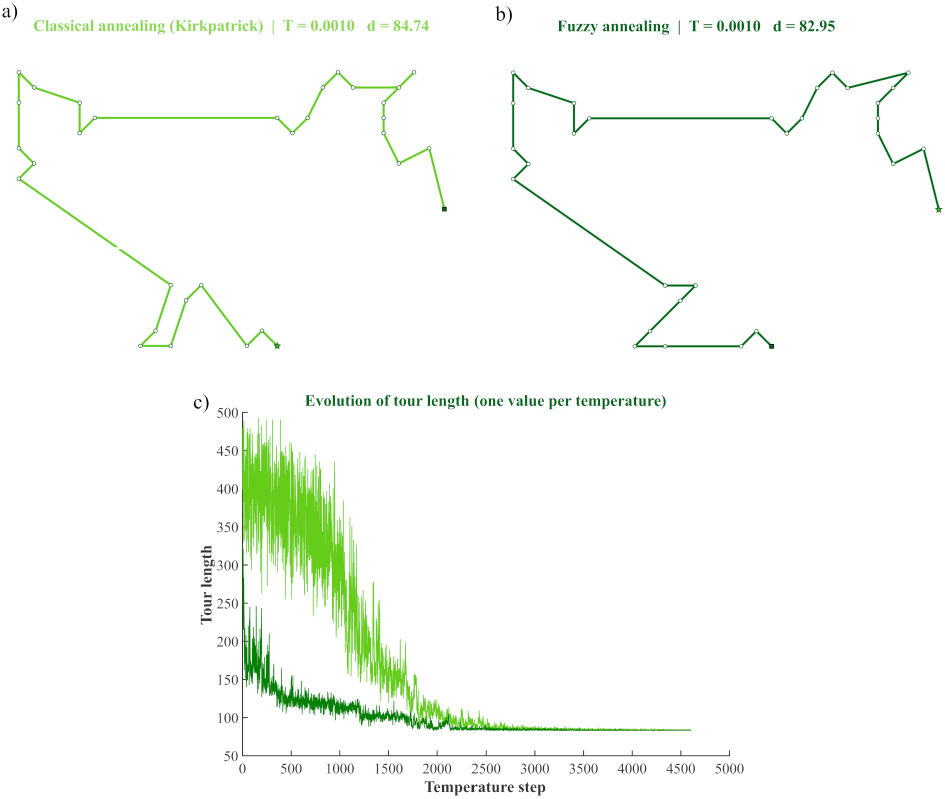}  
	\caption{Comparison of the classical Kirkpatrick simulated annealing (light green) and a fuzzy-logic variant (dark green) applied to the traveling salesman problem. Panels (a) and (b) show the final tours found by the two algorithms: although both reach acceptable solutions, the fuzzy version produces a shorter and more regular path. Panel (c) shows the evolution of tour length over the cooling schedule. The classical method displays the characteristic volatility of probabilistic annealing: large fluctuations, irregular jumps, and repeated excursions into highly suboptimal regions. By contrast, the fuzzy algorithm exhibits a smoother, more disciplined trajectory, in which decreases in tour length are more consistent and inefficient configurations are naturally suppressed by fuzzy membership rules. This illustrates a conceptual difference between probability and fuzzy logic: whereas stochastic annealing treats every trial as potentially acceptable, fuzzy evaluation imposes qualitative judgments (“bad”, “medium”, “good”) that act as epistemic filters, reducing the acceptance of implausible models and guiding the search more rationally through the solution space.}
	\label{Fig:06}
\end{figure}

Figure \ref{Fig:06} offers a detailed comparison between the classical \shortciteN{kirkpatrick1983} simulated annealing algorithm and a fuzzy-logic analogue applied to the traveling salesman problem, revealing not only differences in performance but also deep contrasts in the way uncertainty is treated within each framework. Panels a) and b) show the final tours obtained after cooling. The classical probabilistic scheme (light green), which accepts or rejects trial moves according to the traditional \shortciteN{metropolis1953} rule, converges to a path that is globally coherent but locally irregular, with detours that reflect the algorithm’s willingness to accept poor configurations in the hope of escaping local minima. The fuzzy version (dark green), by contrast, produces a tour that is slightly shorter and structurally smoother. This improvement stems not from a more aggressive optimization strategy but from a fundamentally different treatment of decision-making: candidate solutions are not judged in a binary fashion (“accept” or “reject”), but according to graded membership functions that quantify how “bad”, “medium”, or “good” a tour is. The presence of these qualitative layers biases the search away from implausible or incoherent models even before the algorithm evaluates their numerical merit. Panel c) makes this epistemological contrast explicit by plotting the evolution of the tour length throughout the cooling schedule. The classical method displays a characteristic noise-sawtooth pattern: large amplitude fluctuations, sudden drops followed by dramatic deterioration, and intermittent excursions toward highly suboptimal regions. These oscillations reflect the stochastic heart of probabilistic annealing; random exploration grants freedom to the algorithm but also exposes it to noise and to what might be called “epistemic instability”. At any given time, a poor solution may be accepted simply because the Metropolis criterion, governed by temperature, still authorizes such steps. This behavior is mathematically sound and historically central to the method, yet from an epistemological standpoint it exemplifies a mode of reasoning in which uncertainty is embraced as an engine of exploration. The fuzzy algorithm behaves differently. Its trajectory descends more monotonically, with fluctuations that are both smaller and more structured. The membership functions act as qualitative filters that temper the randomness inherent in the optimization procedure. A candidate tour that is extremely long, corresponding to a model that is clearly unacceptable, receives a near-zero degree of membership in the “good” set, and the algorithm is correspondingly disinclined to accept it, even early in the cooling. Conversely, moderately suboptimal tours may still be considered “acceptable” with a certain degree, allowing the method to retain some of the explorative power of classical annealing while avoiding its most erratic excursions. From this perspective, fuzzy logic does not merely modify the acceptance rule; it reshapes the epistemic landscape of the search by embedding qualitative judgment into the algorithmic process. Seen through this lens, the comparison between probability and fuzzy logic mirrors a broader philosophical dichotomy between two ways of handling uncertainty. Probabilistic annealing relies on randomness as a tool for discovering structure; it navigates by chance, accepting disorder in the short term for the promise of order in the long term. Fuzzy annealing, on the other hand, introduces an intermediate evaluative layer that reflects a kind of “algorithmic common sense”; before randomness can act, candidate models are screened according to linguistic categories that encode coherence, plausibility, or desirability. By doing so, fuzzy logic restricts the exploration to cognitively meaningful regions of the model space, leading to a more stable and directed search. The resulting optimization is not only more efficient but also more interpretable, as each decision can be traced back to explicit qualitative criteria rather than to probabilistic fluctuations alone. In this sense, Figure \ref{Fig:06} does more than compare two numerical methods; it stages a conceptual dialogue between two philosophies of inference. The classical probabilistic approach asserts that global structure can emerge out of local randomness, while the fuzzy approach asserts that randomness itself benefits from being modulated by qualitative judgment. Their juxtaposition makes clear that the logic of optimization, like the logic of scientific reasoning, can be framed either as the aggregation of stochastic trials or as the interaction between numerical evidence and qualitative constraints. The “superiority” of the fuzzy method in this example is therefore not merely computational; it reflects a deeper epistemological insight into how models should be evaluated when uncertainty is present.

Seen in this light, fuzzy logic represents a continuation of the historical movement traced throughout this article. Just as probability extended classical logic by allowing intermediate degrees of belief, fuzzy logic extends rational reasoning by allowing intermediate degrees of meaning. It responds directly to the limitation identified in probabilistic expressivity: the inability to represent vagueness as such. By formalizing graded membership, fuzzy logic enables reason to operate coherently in domains where concepts do not admit sharp boundaries, without collapsing vagueness into randomness.

The emergence of fuzzy logic therefore marks a new stage in the evolution of rationality. After having learned to domesticate chance and to quantify uncertainty, reason now confronts the task of formalizing imprecision itself. This task does not undermine the achievements of probability theory; it complements them. Together, probability and fuzzy logic form a richer epistemic framework, capable of addressing both uncertainty about facts and vagueness of concepts, two dimensions that modern scientific inquiry can no longer afford to conflate.

\section{\label{sec:X}    Geometry without logic. Artificial neural networks and the illusion of model-free knowledge}

The historical trajectory traced in this article has progressively clarified what constitutes rational inference under uncertainty. Probability theory provided a coherent logic for updating beliefs in time, allowing scientific reasoning to integrate new evidence without sacrificing consistency. Fuzzy logic, in turn, addressed a distinct limitation of probabilistic reasoning by formalizing vagueness and graded meaning, making it possible to reason rigorously when concepts themselves lack sharp boundaries. Together, these frameworks articulate a form of rationality that explicitly represents uncertainty, interpretation, and judgment. Against this background, the recent rise of deep learning invites a careful epistemological reassessment.

In what follows, the expression ``neural networks'' refers to artificial neural networks used as computational models, and not to biological neural systems. The discussion below does not address the general philosophical question of whether an embodied artificial system could, in principle, develop forms of understanding, intentionality, or consciousness. It concerns a narrower issue: the epistemic status of present deep-learning architectures when they are used as scientific models for prediction, explanation, and inference. In this restricted sense, the question is not whether artificial intelligence could one day understand the world, but whether current neural architectures make their assumptions, uncertainties, causal structures, and conceptual categories explicit.

Over the past decade, deep artificial neural networks have achieved spectacular empirical success in domains ranging from image recognition to speech processing and time-series prediction. Architectures such as Long Short-Term Memory networks (LSTM; \shortciteNP{hochreiter1997}) and nonlinear autoregressive models with exogenous inputs (NARX; \shortciteNP{leontaritis1985a,leontaritis1985b}) are now routinely employed to model highly complex dynamical systems. In many scientific and engineering contexts, these models appear capable of capturing long-range dependencies, nonlinear interactions, and subtle regularities that were previously inaccessible. This success has fueled the widespread claim that deep learning represents an epistemic rupture: a form of ``model-free'' intelligence able to extract knowledge directly from data, without explicit hypotheses, prior structures, or interpretative frameworks.

Such claims, however, conflate predictive performance with epistemic reasoning. While artificial neural networks undoubtedly extend our capacity to approximate complex input/output relations, their mode of operation differs fundamentally from both probabilistic and fuzzy forms of inference. Unlike Bayesian reasoning, they do not update explicit hypotheses in light of new evidence; unlike fuzzy logic, they do not manipulate graded concepts or qualitative categories. Instead, in their standard use as scientific predictors, they implement a largely geometric form of computation, whose internal coherence is numerical rather than explicitly epistemic.

Mathematically, a deep neural network is a parametric function defined on a high-dimensional vector space. In the now-standard case of ReLU-based architectures (\cf \shortciteNP{householder1941}), each neuron defines a half-space, and each layer composes these half-spaces into a partition of the input space into polytopes of increasing combinatorial complexity (\cf Appendix \ref{App:B}). After training, the network realizes a piecewise-affine mapping whose parameters have been adjusted to minimize a loss function over a dataset. Universality theorems guarantee that, given sufficient depth and width, such networks can approximate any continuous function on a compact domain (\cf Appendix \ref{App:C}). This expressive power, however, should not be mistaken for understanding. Approximation does not entail explanation, and interpolation does not constitute inference.

Sequential architectures such as LSTM or NARX networks illustrate this distinction particularly clearly. These models are often described as possessing ``memory,'' since their outputs depend on past inputs or internal states. Yet this memory is not epistemic in nature. It does not correspond to the updating of beliefs, the revision of hypotheses, or the accumulation of evidence in a logical sense. Rather, it is a state-dependent parameterization of a function that encodes correlations across time. The network does not ask whether a hypothesis remains plausible in light of new data; it simply computes a new output given an updated internal state. Temporal dependence, in this framework, is not equivalent to an arrow of epistemic time (\cf Appendices \ref{App:A} and \ref{App:E}).

\begin{figure}[H]
	\centering
	\includegraphics[width=1\textwidth]{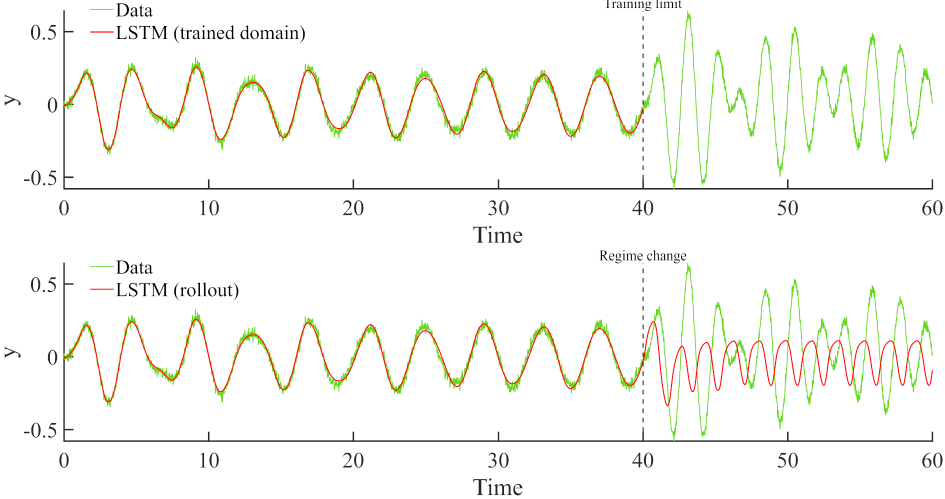}  
	\caption{Interpolation versus extrapolation in an LSTM model. A Long Short-Term Memory (LSTM) network is trained on a stationary regime of a mechanistic dynamical system over the interval $t \in [0,40]$, where $t$ denotes a dimensionless time variable. Top panel: within the training domain, the network achieves an excellent fit (in red) and closely interpolates the observed data (in green). Bottom panel: when the same network is deployed in closed-loop prediction beyond the training interval, under a regime change affecting the external forcing, the predicted signal progressively drifts in phase and amplitude. Although the underlying physical dynamics remain unchanged, the LSTM fails to extrapolate coherently, illustrating the distinction between prediction by interpolation and explanation by invariant mechanisms.}
	\label{Fig:07}
\end{figure}

Figure \ref{Fig:07} provides a concrete numerical illustration of the epistemological distinction developed in this section between interpolation-based prediction and mechanistic explanation. The figure is based on a forced damped oscillator, in which the measured quantity is the scalar displacement $x(t)$. The LSTM is trained only on the first regime, $t \in [0,40]$, where the forcing is stationary. At $t=40$, the governing differential equation is unchanged, but the external forcing changes in amplitude and frequency. The purpose of the experiment is therefore not to show that the physical system changes, but to show that the network has learned the statistical regime of the observations rather than the invariant mechanism generating them.

The upper panel displays the behavior of the LSTM within the training domain. The network is trained exclusively on data (in green) spanning the interval $t \in [0,40]$, during which the system is excited by a stationary forcing. In this regime, the agreement between the observed signal and the LSTM prediction (in red) is remarkably good. The network reproduces both the amplitude and the phase of the oscillations, even in the presence of noise ($\sim 5\%$). At this stage, the model appears to have successfully captured the statistical regularities of the training regime. This visual impression reflects a genuine computational achievement: the LSTM is able to interpolate with high accuracy within the regime represented in the training data.

The lower panel reveals a fundamentally different behavior once the network is asked to operate outside the conditions under which it was trained. Beyond the training limit, the external forcing undergoes a change in amplitude and frequency, while the governing physical law of the system remains unchanged. The LSTM is then run in closed-loop mode, meaning that its own predictions are recursively fed back as inputs for future predictions. Under these conditions, a progressive divergence between the predicted signal and the observed data becomes apparent. The network exhibits a growing phase shift and a loss of amplitude fidelity, ultimately producing oscillations that no longer correspond to the true system response.

This degradation is not the consequence of noise amplification or numerical instability, but rather the manifestation of a deeper limitation. The LSTM has not learned the invariant structure of the dynamical system; instead, it has learned an observational mapping that is valid within the statistical regime represented in the training data. When the input distribution changes, even though the physical mechanism does not, the learned mapping no longer provides reliable guidance. The network continues to generate predictions, but these predictions are no longer constrained by the underlying causal structure of the system.

Figure \ref{Fig:07} thus makes visible a crucial epistemological point: excellent predictive performance within a training domain does not imply explanatory understanding. In the training regime, interpolation suffices, and the distinction between correlation and causation remains latent. Once the regime changes, however, the absence of an explicit representation of dynamical invariants becomes evident. The LSTM's internal memory, often invoked as a surrogate for temporal reasoning, does not encode the system's governing law; it merely stores and propagates patterns extracted from past observations.

By contrasting the apparent success of the network within the training interval with its failure beyond it, Figure \ref{Fig:07} illustrates the limits of model-free learning in a particularly transparent way. The network's behavior underscores the difference between learning a mechanism, which generalizes by invariance, and learning a correlation, which generalizes only by interpolation. This distinction lies at the heart of the argument developed in this section: deep learning methods can be extraordinarily effective predictors under stable conditions, yet remain epistemologically silent with respect to explanation when confronted with changes that require an understanding of underlying structure rather than statistical regularity.

This absence of explicit reasoning has important consequences. Probabilistic inference distinguishes between prior knowledge and observational evidence, and combines them through well-defined logical operations. Fuzzy logic distinguishes between degrees of truth and degrees of membership, allowing qualitative judgment to be formalized. Artificial neural networks, by contrast, tend to collapse heterogeneous sources of information into a single optimization criterion. The loss function does not discriminate, by itself, between uncertainty, vagueness, noise, or conceptual ambiguity; it merely quantifies numerical discrepancy. As a result, the internal representations learned by the network often remain opaque with respect to meaning, causality, and interpretation.

From an epistemological standpoint, this opacity is not accidental but structural. In their standard use as scientific predictors, artificial neural networks are not theories about the world; they are geometric machines for approximating mappings between data spaces. They do not, by themselves, formulate hypotheses, test alternatives, or assign degrees of plausibility to competing explanations. They construct a partition of a vector space in which inputs that are close in a numerical sense tend to produce similar outputs. This mode of operation is highly effective when the task at hand is interpolation within a stable domain of observations. It becomes fragile, however, when confronted with distributional shifts, conceptual changes, or questions that require explicit justification rather than numerical accuracy.

The frequent analogy between deep learning and Laplace's demon is therefore misleading. Where Laplace imagined a deterministic intelligence endowed with complete knowledge of causes, present deep-learning architectures generally operate without explicit causal models. Their apparent determinism can mask a deeper epistemic indifference: they can produce a prediction without representing why that prediction holds. In this respect, deep learning revives, in technological form, an older temptation of scientific reason: to equate regularity with law, correlation with explanation, and performance with understanding.

This analysis does not deny the practical utility of artificial neural networks, nor does it diminish their computational achievements. Rather, it situates them within a broader epistemological landscape. Deep learning excels at capturing correlations in high-dimensional spaces, but, in its usual form, it often does so by bypassing the very structures that probability and fuzzy logic were designed to make explicit: uncertainty, interpretation, and judgment. Its power lies primarily in geometry and optimization, not in explicit logic. The resulting outputs may be accurate, sometimes remarkably so, but they remain epistemically incomplete when they are not accompanied by an explicit account of uncertainty, causality, and conceptual meaning.

The illusion of model-free knowledge thus arises from a category error. Artificial neural networks are not free of models; they embody implicit geometric and statistical assumptions, even when these assumptions are not expressed in the language of hypotheses, priors, likelihoods, or qualitative concepts. What they lack, in their standard use as scientific predictors, is therefore not structure, but an explicit rational structure accessible to interpretation. This statement should not be understood as a claim about the theoretical impossibility of artificial understanding or artificial consciousness. It concerns the epistemic status of present deep-learning architectures when they are used as scientific models. In contrast to probabilistic and fuzzy frameworks, which aim to render uncertainty, inference, and conceptual qualification explicit, deep learning often conceals its assumptions behind layers of numerical optimization. This concealment explains both its empirical power and its epistemological limitations.

In the context of the historical evolution examined in this article, deep learning appears not as the culmination of explicit rationality, but as a powerful yet incomplete detour. It demonstrates how far computation can go when liberated from explicit logic, and at the same time, how much is lost when reasoning is reduced to geometry alone. The challenge for contemporary science is therefore not to oppose artificial neural networks to probabilistic or fuzzy reasoning, but to articulate them within a broader epistemic framework in which uncertainty, vagueness, and causality are once again made explicit.

\section{\label{sec:XI}   Conclusion: reason after probability. Reason after probability}
By retracing the historical development of probability theory from its early combinatorial origins to its contemporary interpretations, this article has argued that probability is not merely a technical apparatus for managing randomness, but a historically evolving form of rationality. Each major transformation of probability theory corresponds to a deep shift in how scientific reason conceives uncertainty, time, and inference. From Pascal’s symmetry of equipossible cases to Bayes and Laplace’s introduction of inductive learning, from Poisson’s temporalization of events to Kolmogorov’s axiomatic closure, probability has progressively incorporated the arrow of time into rational judgment. With Tarantola’s interpretation of probability as a logic of information, this evolution reaches a point of remarkable coherence, where uncertainty, data, and prior knowledge are unified within a single epistemic framework.

Yet this very success makes the internal limits of probabilistic reasoning visible. As shown in this article, probability theory presupposes that the propositions to which it assigns degrees of belief are already well defined. It quantifies uncertainty about facts, but it does not address the imprecision of the concepts through which facts are apprehended. In many contemporary scientific contexts, uncertainty is inseparable from vagueness: models are judged plausible or implausible, signals weak or strong, structures compatible or incompatible with data, without admitting sharp boundaries. Recognizing this limitation does not undermine probability theory; rather, it reveals the need for a complementary extension of rationality.

Fuzzy logic responds to this need by formalizing graded meaning and qualitative judgment. By introducing degrees of membership rather than degrees of belief, it provides a rigorous language for reasoning in domains where concepts are intrinsically imprecise. In this sense, fuzzy logic occupies a position analogous to that once occupied by probability itself: it extends rational reasoning into regions that classical logic and probabilistic calculus cannot fully reach. Probability and fuzzy logic thus address distinct but complementary dimensions of epistemic uncertainty, one concerning the truth of propositions, the other concerning the meaning of the categories involved.

Placed within this historical trajectory, the rise of deep learning acquires a clearer epistemological significance. Neural networks do not represent a further extension of rationality in the sense traced throughout this article, but rather a powerful computational detour. By operating through geometric interpolation in high-dimensional spaces, they achieve impressive predictive performance while bypassing explicit representations of uncertainty, causality, and conceptual judgment. Their success illustrates the effectiveness of correlation-based computation, but also its epistemic opacity. Deep learning excels where interpolation suffices, yet it remains silent with respect to explanation, justification, and meaning.

The contrast is therefore not between old and new technologies, nor between human and artificial intelligence, but between different conceptions of rationality. Probabilistic reasoning introduces coherence and temporal updating; fuzzy logic introduces qualification and graded interpretation; deep learning introduces geometric power without explicit logic. Confusing these modes of reasoning leads to the illusion of model-free knowledge, in which performance is mistaken for understanding and regularity for law.

The broader lesson of this inquiry is that rationality does not progress by accumulation alone, but by differentiation. Each historical extension of reason has clarified what previous frameworks could not express. In this light, the future of scientific reasoning does not lie in the abandonment of probabilistic or fuzzy frameworks, nor in the uncritical celebration of data-driven methods, but in their careful articulation. A mature epistemology of uncertainty must integrate quantitative inference, qualitative judgment, and computational efficiency, without reducing one to the others.

By viewing probability as an evolving form of rationality rather than a fixed mathematical tool, this article offers a perspective from which contemporary debates on artificial intelligence, inference, and explanation can be reassessed. The evolution of probability mirrors the evolution of reason itself: a continuous effort to make uncertainty intelligible, without erasing the complexity of the world it seeks to understand.
\newpage
\bibliographystyle{fchicago}
\bibliography{proba.bib}

@article{barone1978,
  title={A history of the axiomatic formulation of probability from Borel to Kolmogorov: Part I},
  author={Barone, Jack and Novikoff, Albert},
  journal={Archive for history of exact sciences},
  volume={18},
  number={2},
  pages={123--190},
  year={1978},
  publisher={Springer}
}

@article{bayes1763,
  title={An essay towards solving a problem in the doctrine of chances. By the late Rev. Mr. Bayes, FRS communicated by Mr. Price, in a letter to John Canton, AMFR S},
  author={Bayes, Thomas},
  journal={Philosophical transactions of the Royal Society of London},
  number={53},
  pages={370--418},
  year={1763},
  publisher={The Royal Society London}
}

@book{bernoulli1713,
  title={Ars conjectandi, opus posthumum: accedit tractatus de seriebus infinitis, et epistola Gallice scripta de ludo pil{\ae} reticularis},
  author={Bernoulli, Jakob},
  year={1713},
  publisher={Impensis Thurnisiorum Fratrum}
}

@article{borel1909,
  title={Les probabilit{\'e}s d{\'e}nombrables et leurs applications arithm{\'e}tiques},
  author={Borel, {\'E}mile M.},
  journal={Rendiconti del Circolo Matematico di Palermo (1884-1940)},
  volume={27},
  number={1},
  pages={247--271},
  year={1909},
  publisher={Springer Milan Milan}
}

@article{chen2003,
  title={Bayesian filtering: From Kalman filters to particle filters, and beyond},
  author={Chen, Zhe and others},
  journal={Statistics},
  volume={182},
  number={1},
  pages={1--69},
  year={2003}
}

@book{condorcet1785,
  title={Essai sur l'application de l'analyse {\`a} la probabilit{\'e} des d{\'e}cisions rendues {\`a} la pluralit{\'e} des voix},
  author={Condorcet, Nicolas},
  year={1785},
  publisher={de l'Imprimerie royale, Paris, France}
}

@book{cox1961,
  title={The algebra of probable inference},
  author={Cox, Richard T},
  year={1961},
  publisher={Johns Hopkins University Press}
}

@article{cybenko1989,
  title={Approximation by superpositions of a sigmoidal function},
  author={Cybenko, George},
  journal={Mathematics of control, signals and systems},
  volume={2},
  number={4},
  pages={303--314},
  year={1989},
  publisher={Springer}
}

@incollection{daston1992,
  title={The doctrine of chances without chance: determinism, mathematical probability, and quantification in the seventeenth century},
  author={Daston, Lorraine J},
  booktitle={The Invention of Physical Science: Intersections of Mathematics, Theology and Natural Philosophy Since the Seventeenth Century Essays in Honor of Erwin N. Hiebert},
  pages={27--50},
  year={1992},
  publisher={Springer}
}

@article{edwards1963,
  title={Bayesian statistical inference for psychological research.},
  author={Edwards, Ward and Lindman, Harold and Savage, Leonard J},
  journal={Psychological review},
  volume={70},
  number={3},
  pages={193},
  year={1963},
  publisher={American Psychological Association}
}

@inproceedings{finetti1937,
  title={La pr{\'e}vision: ses lois logiques, ses sources subjectives},
  author={De Finetti, Bruno},
  booktitle={Annales de l'institut Henri Poincar{\'e}},
  volume={7},
  number={1},
  pages={1--68},
  year={1937}
}

@incollection{galton1814,
  title={On the probability of the extinction of families},
  author={Galton, Francis and Watson, Henry William},
  booktitle={Mathematical Demography: Selected Papers},
  pages={399--406},
  year={1814},
  publisher={Springer}
}

@book{gauss1809,
  title={Theoria motus corporum coelestium in sectionibus conicis solem ambientium auctore Carolo Friderico Gauss},
  author={Gauss, Carl Friedrich},
  publisher={Frid. Perthes et IH Besser},
  year={1809}
}

@article{gibert2024,
  title={Information theory, Signal Analysis and Inverse Problem},
  author={Gibert, Dominique and Lopes, Fernando and Courtillot, Vincent and Mou{\"e}l, Jean-Louis Le and Boul{\'e}, Jean-Baptiste},
  journal={arXiv preprint arXiv:2408.16361},
  year={2024}
}

@book{hald2005,
  title={A history of probability and statistics and their applications before 1750},
  author={Hald, Anders},
  year={2005},
  publisher={John Wiley \& Sons}
}

@article{hochreiter1997,
  title={Long short-term memory},
  author={Hochreiter, Sepp and Schmidhuber, J{\"u}rgen},
  journal={Neural computation},
  volume={9},
  number={8},
  pages={1735--1780},
  year={1997},
  publisher={MIT press}
}

@article{hornik1991,
  title={Approximation capabilities of multilayer feedforward networks},
  author={Hornik, Kurt},
  journal={Neural networks},
  volume={4},
  number={2},
  pages={251--257},
  year={1991},
  publisher={Elsevier}
}

@article{householder1941,
  title={A theory of steady-state activity in nerve-fiber networks: I. Definitions and preliminary lemmas},
  author={Householder, Alston S},
  journal={The bulletin of mathematical biophysics},
  volume={3},
  number={2},
  pages={63--69},
  year={1941},
  publisher={Springer}
}

@book{huygens1657,
  title={Exercitationum Mathematicarum Liber V, De Ratiociniis in Ludo Aleae},
  author={Huygens, Christiaan},
  year={1657},
  pages={pp 519-534},
  publisher={Ex officinia J. Elsevirii}
}

@book{jaynes2003,
  title={Probability theory: The logic of science},
  author={Jaynes, Edwin T},
  year={2003},
  publisher={Cambridge university press}
}

@book{jeffreys1939,
  title={Theory of probability},
  author={Jeffreys, H},
  year={1939},
  publisher={Oxford Univ. Press}
}

@article{kalman1960,
  title={A new approach to linear filtering and prediction problems},
  author={Kalman, Rudolph Emil},
  year={1960}
}

@article{kirkpatrick1983,
  title={Optimization by simulated annealing},
  author={Kirkpatrick, Scott and Gelatt Jr, C Daniel and Vecchi, Mario P},
  journal={science},
  volume={220},
  number={4598},
  pages={671--680},
  year={1983},
  publisher={American association for the advancement of science}
}

@article{kolmogorov1933,
  title={Grundbegriffe der wahrscheinlichkeitsrechnung},
  author={Kolmogorov, Andrey},
  year={1933},
  publisher={Springer}
}

@article{laplace1774,
  title={M{\'e}moire sur la probabilit{\'e} de causes par les {\'e}venements},
  author={Laplace, Pierre Simon},
  journal={M{\'e}moire de l'acad{\'e}mie royale des sciences},
  year={1774}
}

@book{laplace1812,
  title={Th{\'e}orie analytique des probabilit{\'e}s},
  author={Laplace, Pierre Simon},
  volume={7},
  year={1812},
  publisher={Courcier}
}

@book{laplace1814,
  title={Essai philosophique sur les Prababilités},
  author={Laplace, Pierre Simon},
  year={1814},
  publisher={Courcier}
}

@book{legendre1806,
  title={Nouvelles m{\'e}thodes pour la d{\'e}termination des orbites des com{\`e}tes: avec un suppl{\'e}ment contenant divers perfectionnemens de ces m{\'e}thodes et leur application aux deux com{\`e}tes de 1805},
  author={Legendre, Adrien Marie},
  year={1806},
  publisher={Courcier}
}

@article{leontaritis1985a,
  title={Input-output parametric models for non-linear systems part I: deterministic non-linear systems},
  author={Leontaritis, I Jꎬ and Billings, Stephen A},
  journal={International journal of control},
  volume={41},
  number={2},
  pages={303--328},
  year={1985},
  publisher={Taylor \& Francis}
}

@article{leontaritis1985b,
  title={Input-output parametric models for non-linear systems part II: stochastic non-linear systems},
  author={Leontaritis, IJ and Billings, Steve A},
  journal={International journal of control},
  volume={41},
  number={2},
  pages={329--344},
  year={1985},
  publisher={Taylor \& Francis}
}

@book{lindley1972,
  title={Bayesian statistics: A review},
  author={Lindley, Dennis Victor},
  year={1972},
  publisher={SIAM}
}

@book{lyraud2003,
  editor    = {Lyraud, Pierre and Plazenet, Laurence},
  title     = {L'Oeuvre de Blaise Pascal},
  year      = {2003},
  publisher = {Éditions du 400e anniversaire},
  address   = {France},
  series    = {Collection Mollat},
  language  = {french}
}

@article{metropolis1953,
  title={Equation of state calculations by fast computing machines},
  author={Metropolis, Nicholas and Rosenbluth, Arianna W and Rosenbluth, Marshall N and Teller, Augusta H and Teller, Edward},
  journal={The journal of chemical physics},
  volume={21},
  number={6},
  pages={1087--1092},
  year={1953},
  publisher={American Institute of Physics}
}

@book{moivre1718,
  author    = {de Moivre, Abraham},
  title     = {The Doctrine of Chances or a Method of calculating the probability of events in play},
  year      = {1718},
  publisher = {W. Pearson},
  address   = {London},
  language  = {english}
}

@article{montufar2014,
  title={On the number of linear regions of deep neural networks},
  author={Mont{\'u}far, Guido and Pascanu, Razvan and Cho, Kyunghyun and Bengio, Yoshua},
  journal={Advances in neural information processing systems},
  volume={27},
  year={2014}
}

@article{mosegaard2011,
  title={Quest for consistency, symmetry, and simplicity—The legacy of Albert Tarantola},
  author={Mosegaard, Klaus},
  journal={Geophysics},
  volume={76},
  number={5},
  pages={W51--W61},
  year={2011},
  publisher={Society of Exploration Geophysicists}
}

@article{neyman1933,
  title={On the problem of the most efficient tests of statistical},
  author={Neyman, J and Pearson, ES},
  journal={London: Philosophical Transactions of the Royal Society of London},
  year={1933}
}

@article{neyman1977,
  title={Frequentist probability and frequentist statistics},
  author={Neyman, Jerzy},
  journal={Synthese},
  volume={36},
  number={1},
  pages={97--131},
  year={1977},
  publisher={JSTOR}
}

@article{ore1960,
  title={Pascal and the invention of probability theory},
  author={Ore, Oystein},
  journal={The American Mathematical Monthly},
  volume={67},
  number={5},
  pages={409--419},
  year={1960},
  publisher={Taylor \& Francis}
}

@book{pascal1665,
  title={Trait{\'e} du triangle arithmetique: avec quelques autres petits traitez sur la mesme matiere},
  author={Pascal, Blaise},
  year={1665},
  publisher={Chez Guillaume Desprez}
}

@book{poisson1837,
  title={Recherches sur la probabilit{\'e} des jugements en mati{\`e}re criminelle et en mati{\`e}re civile: pr{\'e}c{\'e}d{\'e}es des r{\`e}gles g{\'e}n{\'e}rales du calcul des probabilit{\'e}s},
  author={Poisson, Sim{\'e}on-Denis},
  year={1837},
  publisher={Bachelier}
}

@book{ramsey1931,
  title={Foundations of mathematics and other logical essays},
  author={Ramsey, Frank Plumpton},
  year={1931},
  publisher={Routledge}
}

@article{reeves2015,
  title={The secularization of chance: Toward understanding the impact of the probability revolution on Christian belief in divine providence},
  author={Reeves, Josh},
  journal={Zygon{\textregistered}},
  volume={50},
  number={3},
  pages={604--620},
  year={2015},
  publisher={Wiley Online Library}
}

@book{tarantola2005,
  title={Inverse problem theory and methods for model parameter estimation},
  author={Tarantola, Albert},
  year={2005},
  publisher={SIAM}
}

@book{savage1954,
  title={The foundations of statistics},
  author={Savage, Leonard J},
  year={1954},
  publisher={Courier Corporation}
}

@article{tarantola1982,
  title={Inverse problems= quest for information},
  author={Tarantola, Albert and Valette, Bernard and others},
  journal={Journal of geophysics},
  volume={50},
  number={1},
  pages={159--170},
  year={1982}
}

@article{venn1866,
  title={The Logic of Chance.},
  author={Venn, J},
  journal={New York: Chelsea},
  year={1866}
}

@article{yule1925,
  title={II—A mathematical theory of evolution, based on the conclusions of Dr. JC Willis, FR S},
  author={Yule, George Udny},
  journal={Philosophical transactions of the Royal Society of London. Series B, containing papers of a biological character},
  volume={213},
  number={402-410},
  pages={21--87},
  year={1925},
  publisher={The Royal Society London}
}

@article{zadeh1965,
  title={Fuzzy sets},
  author={Zadeh, Lotfi A},
  journal={Information and control},
  volume={8},
  number={3},
  pages={338--353},
  year={1965},
  publisher={Elsevier}
}

@article{zadeh1978,
  title={Fuzzy sets as a basis for a theory of possibility},
  author={Zadeh, Lotfi Asker},
  journal={Fuzzy sets and systems},
  volume={1},
  number={1},
  pages={3--28},
  year={1978},
  publisher={Elsevier}
}
\newpage

\counterwithin{equation}{section}        
\renewcommand{\theequation}{%
	\thesection.\ifnum\value{equation}<10 0\fi\arabic{equation}%
}

\appendix
\titleformat{\section}
{\normalfont\Large\bfseries}       
{Appendix~\thesection}             
{1em}                              
{}                 

\setcounter{figure}{0}
\renewcommand{\thefigure}{A\padzeroes[2]{\arabic{figure}}}
\section{ Correlations and convolutions: symmetry of time vs. arrow of time\label{App:A}}
	To better understand some of the concepts discussed above, it is instructive to make a more technical digression on two mathematical operations closely related to probability: correlation and convolution. These two notions offer a vivid analogy for the idea of temporal symmetry versus the arrow of time, which we have already mentioned in passing. In probability theory as in signal processing, a correlation measures the degree of association or similarity between two random variables, or two signals, without implying any direction of causality. Mathematically, the correlation between two variables $X$ and $Y$ can be quantified by the Pearson correlation coefficient,
	\begin{equation}
		\rho_{X,Y} = \dfrac{Cov(X,Y)}{\sigma_{X}\sigma_{Y}},
	\end{equation}

	where, $\mathrm{Cov}(X,Y) = E[(X-E[X])(Y-E[Y])]$ is the covariance. This quantity is symmetric, \ie $\rho_{XY} = \rho_{YX}$. More generally, the cross-correlation function between two signals $x(t)$ and $y(t)$ is defined, in signal processing, as,
	\begin{equation}
		(x \star y)(\tau) = \int_{-\infty}^{+\infty} x(t)\,y(t+\tau)\,dt
	\end{equation} 
	
	which is essentially a convolution product without time reversal of one of the signals. In the particular case of the autocorrelation of a stationary signal $x(t)$, one obtains a function $R_{xx}(\tau)$ hat depends on the time lag $\tau$ and satisfies $R_{xx}(-\tau)=R_{xx}(\tau)$. The autocorrelation is an even (ie. symmetric) function of $\tau$. Intuitively, this means that the similarity of $x(t)$ with itself shifted by $+\tau$ is the same as when shifted by $-\tau$, only the degree of similarity matters, not the temporal order. In terms of reasoning, correlation reveals a mutual relationship between two phenomena without indicating which is the cause and which the effect. This is why one often says that “correlation does not imply causation”; correlations are symmetric and timeless. They may reflect underlying symmetries of the system, \eg two variables influenced by the same hidden factor, or mere statistical coincidence, but to remain at the level of correlation is not yet to introduce a causal arrow.
	
	By contrast, convolution is an operation that, when used to combine an input signal with the impulse response of a system, introduces an explicit temporal orientation. The convolution of two functions $f$ and $g$ is defined as, 
	\begin{equation}
			(f * g) (t) = \int_{-\infty}^{+\infty} f(u) g(t-u) du.
	\end{equation}

	Mathematically, convolution is also commutative $f\star g=g \star f$, but when $f$ is interpreted as a cause, an input signal, and $g$ as a response kernel, the system, one usually imposes $g(t)=0 \quad \forall t<0$, the condition of physical causality, meaning that the system cannot respond before the signal is applied. In this case, the formula reduces to,
	\begin{equation}
		(f * g)(t) = \int_{0}^{t} f(u)\,g(t-u)\,du \quad \forall t\ge0,
	\end{equation}
	
	which shows that the output at time  depends on the past values of the input, \ie. from 0 to $t$, but not on its future values. This expresses an arrow of time; the direction runs from cause to effect, from past to present. Every causal convolution thus encodes a form of practical irreversibility. In principle, given $g$ and the output, one can retrieve the input by deconvolution, but this operation is often nontrivial and highly sensitive to perturbations, a consequence of the information loss inherent in the mixing performed by convolution.

	In probability theory, convolution appears in several temporal contexts. For example, the distribution of the sum of two independent random variables is the convolution of their individual distributions: if $X$ and $Y$ are independent, the density of $Z = X + Y$ is,
	\begin{equation}
		(f_{X} * f_{Y})(z) = \int f_{X}(x)\,f_{Y}(z-x)\,dx.
	\end{equation}
	Although $X+Y=Y+X$ , a symmetry, this calculation can be interpreted as the sequential aggregation of two random contributions. Similarly, in Markov processes, the evolution of the probability distribution follows the Chapman-Kolmogorov equation, which is an integral convolution; the distribution after two steps is the convolution of the distribution after one step with the transition distribution of the next step. This formalizes the idea that the future is constructed from the present through a probabilistic transition kernel. This construction has an intrinsic irreversibility, except in special cases of reversible processes, because it is built upon a given temporal orientation, the composition of transitions in the direction of time.
	
	To fix ideas, consider the following analogy; correlation is like comparing two books to see whether they resemble each other, same words, same chapters, without worrying about who might have copied from whom, whereas convolution is like reading a book sequentially to see how the story unfolds chapter by chapter. Correlation detects a global symmetry or similarity, regardless of order, while convolution represents an ordered accumulation, where each chapter builds upon the previous one. In science, identifying a correlation is often the starting point, we observe that two phenomena vary together, but explaining that correlation requires the introduction of a causal model, which, in essence, is of the convolution type, \eg phenomenon $A$ influencing $B$ over time.
	
	Thus, returning to probability theory and to reason, one might say that reasoning by correlation corresponds to the phase in which we gather evidence and observe symmetric relationships, \eg. two correlated symptoms of a disease,  a form of static associative reasoning. Reasoning by convolution, or through a dynamic model, goes a step further by introducing a sequential structure; one imagines a mechanism that produces these correlations in a given temporal direction, \eg. the progression of the disease first causing symptom $A$ and then symptom $B$. The sequential Bayesian inference discussed earlier,  the step-by-step updating of probability, is precisely convolutive in nature; prior knowledge is combined with new information to yield posterior knowledge, and this process unfolds recursively. Each update acts as a small convolutional fold, \ie a multiplication by the normalized likelihood, that integrates the effect of new data.
	
	In terms of temporal symmetry, one can say that correlation preserves a past-future symmetry, it would remain the same if time were reversed in a stationary system, whereas convolution breaks this symmetry by introducing a preferred direction. This reflects a fundamental feature of modern reason; to understand the world, we no longer limit ourselves to identifying symmetries and static structures; we also seek directed laws of evolution. We accept that to know something is also to know how it changes and how it influences other things over time. Probability, too, has had to integrate this dimension, from an originally static theory,  Laplace could calculate \apriori probabilities of causes or events outside of time, it has become a dynamic theory. Today, the Kalman filter or dynamic Bayesian networks are standard tools for tracking systems in evolution.
	
	In conclusion to this section, the opposition between correlation and convolution is nothing less than a mathematical metaphor for the evolution of probabilistic thought: from the static to the dynamic, from simple association to causality, from timeless symmetry to temporal orientation. Probability theory provides the language to quantify and manipulate both correlations, measuring the joint entropy of variables and their symmetric interdependencies, and convolutions, combining distributions throughout a process. It thus enables reason to address both problems of static understanding, what is the structure of relationships between variables ?  and problems of evolution, how do knowledge and systems change over time?

\newpage 
\setcounter{figure}{0}
\renewcommand{\thefigure}{B\padzeroes[2]{\arabic{figure}}}
\section{Viewing the deep network as a piecewise affine map\label{App:B}}	
Let us consider a ReLU activated network of depth $L$, producing a mapping,
\begin{equation}
	f: \mathbb{R}^{d} \rightarrow  \mathbb{R}^{k},
\end{equation}

the ReLU nonlinearity is defined by,
\begin{equation}
		\textrm{ReLU}(x) = max(0,x).
\end{equation}

A central property of modern deep learning is therefore (\eg  \shortciteNP{montufar2014}): any ReLU network implements a piecewise-affine mapping, that is,
\begin{equation}
	f(x) = \mathcal{A}_{i}(x) + b_{i}, \quad \forall x \in  \mathcal{R}_{i}
\end{equation}

where the $\mathcal{R}_{i}$ are convex polytopes forming a partition of the input space. This structure is crucial for understanding the nature of neural networks; the network learns no law, it extracts no causality, it infers no probability, it approximates a function by juxtaposing a large number of affine planes. Depth plays an exponential role here. The maximal number of affine regions $\mathcal{R}_{i}$ realizable by a network of width $n$ and depth $L$ satisfies,
\begin{equation}
	N_{regions} (n,L) \ge \prod_{l=1}^{L} \sum_{j=0}^{d} \left( \begin{array}{c} n_{l} \\ j \end{array}\right)
\end{equation}

which explodes as soon as $L$ exceeds 5 or 6 layers. In other words, a deep network constructs a fractal geometry of polytopes, capable of representing virtually any shape, yet without ever reaching the very notion of a model.

\newpage 
\setcounter{figure}{0}
\renewcommand{\thefigure}{B\padzeroes[2]{\arabic{figure}}}
\section{Strong interpolation, nonexistent extrapolation\label{App:C}}	
By construction, the function learned by a network within the training domain behaves as a very high-dimensional interpolation. The universality of neural networks can be formulated precisely as follows (\eg \shortciteNP{cybenko1989,hornik1991}), for any continuous function  defined on a compact set $K \subset \mathbb{R}^{d}$ and any $\varepsilon > 0$, there exists a network  such that,
\begin{equation}
	\underset{x \in K}{\textrm{supp}}|f(x)-g(x)|<\varepsilon
\end{equation}

But this theorem applies only to a compact set $K$, with no temporal structure, no notion of cause, and no extrapolation beyond $K$. Thus, its essential implication is negative, a network knows nothing outside the support of the data. The function obtained outside the training domain is often arbitrary, linear, because no new affine region is activated, or on the contrary divergent, because the matrix stack amplifies directions that were never tested. The machine excels at reproducing what is known, yet remains blind to what has never been seen. This is the opposite of probabilistic rationality, where a law entails out-of-sample predictions.

\newpage 
\setcounter{figure}{0}
\renewcommand{\thefigure}{B\padzeroes[2]{\arabic{figure}}}
\section{Absence of internal temporality: contrast with Bayes and Tarantola\label{App:D}}	
Throughout the probabilistic tradition, from Bayes to Kolmogorov, and later Tarantola, the inference process is written,
\begin{equation}
	p(\theta,d) \propto p(d,\theta) p(\theta),
\end{equation}

and it is the observation that shapes the likelihood. This dynamic is intrinsically temporal; it possesses an epistemological arrow of time. Nothing comparable exists in deep learning. Training a network simply amounts to minimizing a cost function,
\begin{equation}
	\underset{W}{\textrm{min}}\dfrac{1}{N} \sum_{i=1}^{N}l(f_{W}(x_{i}),y_{i}),
\end{equation}

where $W$ denotes the collection of weights. Gradient descent does nothing conceptually: it aggregates no information, and has no internal logic akin to the probabilistic “AND” introduced by Tarantola, 
\begin{equation*}
	content...\textrm{posterior} \sim \underbrace{\textrm{prior}}_{\textrm{pre-existing information}}  \textrm{AND} \underbrace{\textrm{data}}_{\textrm{new information}} 
\end{equation*}

The network knows neither hypotheses, nor uncertainties, nor likelihoods. It optimizes a score.

\newpage 
\setcounter{figure}{0}
\renewcommand{\thefigure}{B\padzeroes[2]{\arabic{figure}}}
\section{Structural confusion between correlation and causation\label{App:E}}	
Since the learned function is an interpolation in $\mathbb{R}^{d}$, any deep network in fact performs geometric correlations learned locally in the data. No internal operation is capable of distinguishing an accidental regularity, from a structural law, from a causal mechanism. Mathematically, a deep network implements a function of the form,
\begin{equation}
	f(x) = \sum_{i} \alpha_{i} \mathbb{I}_{\mathcal{R}_{i}} (x) (\mathcal{A}_{i}x + b_{i}), 
\end{equation}

that is, a combination of planes with no internal syntax. There is no hypothesis space, no competition between causes, no Bayesian weighting, $P(H_{i}|d )$ no such thing exists. In emblematic real-world applications (Google Flu Trends, trading systems, medical imaging), this absence of internal causality leads to abrupt collapses as soon as the data distribution shifts (concept drift). The machine has not failed: it had never inferred a law in the first place.

\end{document}